\documentclass{clv3}

\NewCommandCopy{\cnumdef}{\numdef}
\NewCommandCopy{\endcnumdef}{\endnumdef}
\let\numdef\relax \let\endnumdef\relax

\usepackage{pgfplots}
\usepackage{color}
\usepackage{booktabs}
\usepackage{xcolor}

\usepackage{changes}
\usepackage{CJKutf8}
\usepackage{hyperref}
\definecolor{darkblue}{rgb}{0, 0, 0.5}
\hypersetup{colorlinks=true,citecolor=darkblue, linkcolor=darkblue, urlcolor=darkblue}
\usepackage{forest}
\usepackage{amsmath}
\usepackage{everyshi}
\usepackage{multirow}
\usepackage{multicol}
\usepackage{lipsum}
\pgfplotsset{compat=1.18}

\begin{document}
\issue{x}{y}{2023}


\runningtitle{A Survey on LLM-Generated Text Detection}

\runningauthor{Wu et al.}


\title{A Survey on LLM-Generated Text Detection: Necessity, Methods, and Future Directions}

\author{Junchao Wu}
\affil{NLP$^2$CT Lab, Faculty of Science and Technology\\
Institute of Collaborative Innovation\\
University of Macau\\
{\tt nlp2ct.junchao@gmail.com}}

\author{Shu Yang}
\affil{NLP$^2$CT Lab, Faculty of Science and Technology\\
Institute of Collaborative Innovation\\
University of Macau\\
{\tt nlp2ct.shuyang@gmail.com}}

\author{Runzhe Zhan}
\affil{NLP$^2$CT Lab, Faculty of Science and Technology\\
Institute of Collaborative Innovation\\
University of Macau\\
{\tt nlp2ct.runzhe@gmail.com}}

\author{Yulin Yuan$^*$}
\affil{Department of Chinese Language and Literature, Faculty of Arts and Humanties\\
University of Macau\\
{\tt yulinyuan@um.edu.mo} \\
Department of Chinese Language and Literature, Faculty of Humanities\\
Peking University\\
{\tt yuanyl@pku.edu.cn}}

\author{Derek Fai Wong\thanks{Yulin Yuan and Derek Fai Wong are co-coresponding authors.}}
\affil{NLP$^2$CT Lab, Faculty of Science and Technology\\
Institute of Collaborative Innovation\\
University of Macau\\
{\tt derekfw@um.edu.mo}}

\author{Lidia Sam Chao}
\affil{NLP$^2$CT Lab, Faculty of Science and Technology\\
State Key Laboratory of Internet of Things for Smart City\\
University of Macau\\
{\tt lidiasc@um.edu.mo}}

\maketitle

\begin{abstract}

The powerful ability to understand, follow, and generate complex language emerging from large language models (LLMs) makes \textbf{LLM-generated text} flood many areas of our daily lives at an incredible speed and is widely accepted by humans. As LLMs continue to expand, there is an imperative need to develop detectors that can detect LLM-generated text. This is crucial to mitigate potential misuse of LLMs and safeguard realms like artistic expression and social networks from harmful influence of LLM-generated content. The \textbf{LLM-generated text detection} aims to discern if a piece of text was produced by an LLM, which is essentially a binary classification task. The detector techniques have witnessed notable advancements recently, propelled by innovations in watermarking techniques, statistics-based detectors, neural-base detectors, and human-assisted methods. In this survey, we collate recent research breakthroughs in this area and underscore the pressing need to bolster detector research. We also delve into prevalent datasets, elucidating their limitations and developmental requirements. Furthermore, we analyze various LLM-generated text detection paradigms, shedding light on challenges like out-of-distribution problems, potential attacks, real-world data issues and the lack of effective evaluation framework. Conclusively, we highlight interesting directions for future research in LLM-generated text detection to advance the implementation of responsible artificial intelligence (AI). Our aim with this survey is to provide a clear and comprehensive introduction for newcomers while also offering seasoned researchers a valuable update in the field of LLM-generated text detection. The useful resources are publicly available at: \url{https://github.com/NLP2CT/LLM-generated-Text-Detection}.

\end{abstract}

\section{Introduction}
With the rapid development of LLMs, the text generation capabilities of LLMs have reached a level comparable to human writing~\cite{openai2023gpt4,claudemodelcard,chowdhery2022palm}. LLMs have permeated various aspects of daily life and play a vital role become pivotal in many professional workflows \cite{veselovsky2023artificial}, facilitating tasks such as advertising slogan creation \cite{murakami2023natural}, news composition \cite{yanagi2020fake}, story generation \cite{yuan2022wordcraft}, and code generation \cite{becker2023programming,zheng2023codegeex}. \textcolor{black}{A recent research from \citet{DBLP:journals/corr/abs-2305-09820} indicates that the relative quantity of AI-generated news articles on mainstream websites has risen by 55.4\%, whereas on websites known for disseminating misinformation, it has risen by 457\% from January 1, 2022, to May 1, 2023.} Furthermore, their impact significantly shapes the progression of numerous sectors and disciplines, including education \cite{susnjak2022chatgpt}, law \cite{cui2023chatlaw}, biology \cite{piccolo2023many}, and medicine \cite{thirunavukarasu2023large}. 

The powerful generation capabilities of LLMs have rendered it challenging for individuals to discern between LLM-generated and human-written texts, resulting in the emergence of intricate concerns. The concerns regarding LLM-generated text originate from two perspectives. Firstly, LLMs are susceptible to fabrications \cite{ji2023survey}, reliance on outdated information, and heightened sensitivity to prompts. These vulnerabilities can facilitate the spread of erroneous knowledge \cite{christian2023cnet}, undermine technical expertise \cite{rodriguez2022cross,aliman2021epistemic}, and promote plagiarism \cite{lee2023language}. Secondly, there exists the risk of malicious exploitation of LLMs in activities such as disinformation dissemination \cite{pagnoni2022threat,lin2021truthfulqa}, online fraudulent schemes \cite{weidinger2021ethical,ayoobi2023looming}, social media spam production \cite{mirsky2022threat}, and academic dishonesty, especially with students employing LLMs for essay writing \cite{stokel2022ai,kasneci2023chatgpt}. Concurrently, LLMs increasingly shoulder the data generation responsibility in AI research, leading to the recursive use of LLM-generated text in their own training and assessment. A recent analysis, titled Model Autophagy Disorder (MAD) \citep{alemohammad2023selfconsuming}, raised alarms over this AI data feedback loop. As generative models undergo iterative improvements, LLM-generated text may gradually replace the need for human-curated training data. This could potentially lead to a reduction in the quality and diversity of subsequent models. In essence, the consequences of LLM-generated text encompass both societal \cite{cardenuto2023age} and academic \cite{yu2023cheat} risks, and the use of LLM-generated data will hinder the future development of LLMs and detection technology. 

However, for the LLM-generated text detection task, \textcolor{black}{current detection technologies, including the discriminatory capabilities ~\cite{price2023effectiveness} of commercial detectors are unreliable. They are primarily biased towards classifying outputs as human-written text, rather than detecting text generated by LLMs \cite{walters2023effectiveness,weber2023testing,DBLP:journals/corr/abs-2306-15666}.} Detection methods that rely on human are also unreliable and have very low accuracy, even only slightly better than random classification~\cite{uchendu2021turingbench,dou2021gpt,clark2021all,DBLP:journals/corr/abs-2303-17650,Soni2023ComparingAS}. Furthermore, the ability of humans to identify LLM-generated text is often lower than that of detectors or detection algorithms in various environmental settings \cite{ippolito2019automatic,Soni2023ComparingAS}. Thus, there is an imperative demand for robust detectors to identify LLM-generated text effectively. Establishing such mechanisms is pivotal to mitigating LLM misuse risks and fostering responsible AI governance in the LLM era \cite{stokel2023chatgpt,porsdam2023generative,shevlane2023model}.

Research on the detection of LLM-generated text has received lots of attention even before the advent of ChatGPT, especially in areas such as early identification of deepfake text \cite{pu2023deepfake}, machine-generated text detection \cite{jawahar2020automatic} and \textcolor{black}{authorship attribution~\cite{aa}}. Typically, this problem was regarded as a classification task, discerning between LLM-generated text and human-written text \cite{jawahar2020automatic}. Back to this research stage, the detection task was predominantly focusing on translation-generated texts and utilizing \textcolor{black}{simple} statistical methods. The introduction of ChatGPT has sparked a significant surge in interest surrounding LLMs, heralding a paradigm shift in the research landscape. In response to the escalating challenges posed by LLM-generated text, the NLP community has intensely pursued solutions, delving into LLM-generated text detection and related attacking methodologies. While \citet{crothers2023machine, tang2023science} recently presented reviews on this topic, we argue that its depth of detection methods is insufficient \textcolor{black}{(We discuss related work in detail in \autoref{related_works})}. 

In this article, we furnish a meticulous and profound review of contemporary research on LLM-generated text detection, aiming to guide researchers through the challenges and prospective research trajectories. We investigate the latest breakthroughs, beginning with an introduction to the task of LLM-generated text detection, the underlying mechanisms of text generation by LLMs, and the sources of LLM's enhanced text generation capabilities. We also shed light on the contexts and imperatives of LLM-generated text detection. Furthermore, we spotlight popular datasets and benchmarks for the task, exposing their current deficiencies to stimulate the creation of more refined \textcolor{black}{data resources}. Our discussion extends to the latest detector studies. In addition to the traditional neural-based methods and statistical-based methods, we also report watermarking techniques, and human-assisted methods. A subsequent analysis pinpoints research limitations in LLM-generated text detectors, highlighting critical areas like out-of-distribution challenges, potential attacks, \textcolor{black}{real-world data issues, and the lack of effective evaluation framework}. Conclusively, we ponder upon potential directions for future research, aiming to help the development of efficient detectors. 
\section{Background}

\subsection{LLM-generated Text Detection Task}

\begin{figure*}[ht]
\centering
\includegraphics[width=0.9\textwidth, trim=0 0 0 0]{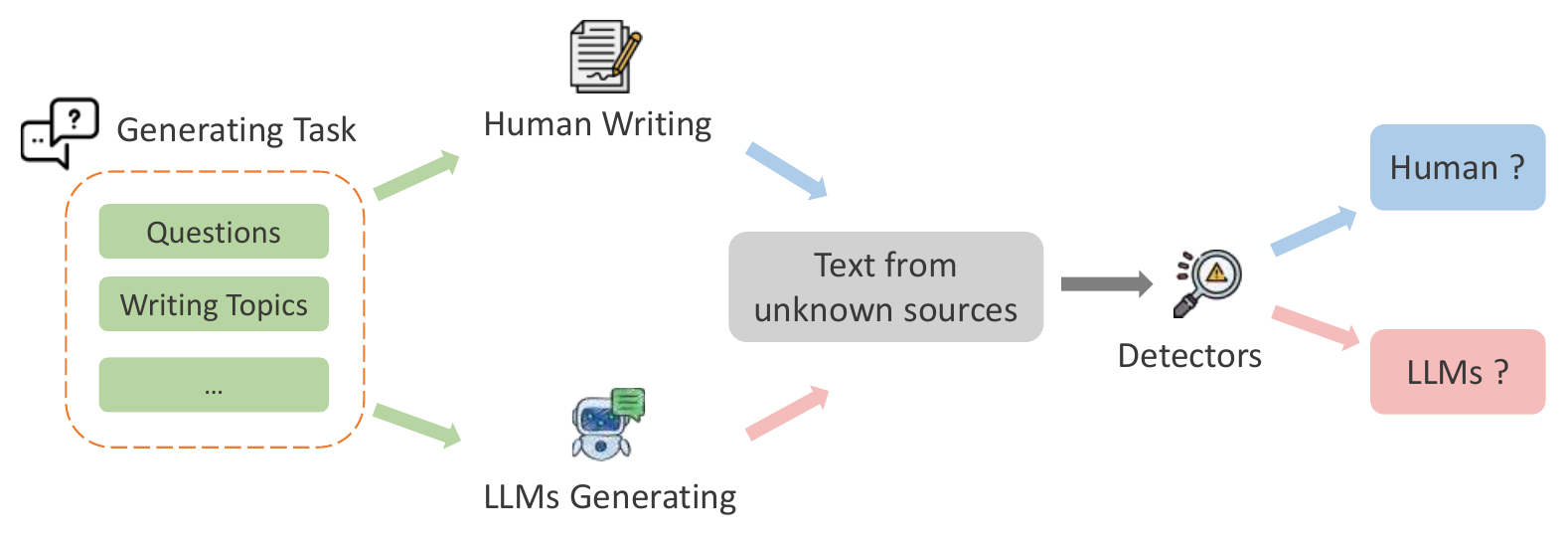}
\caption{Toy picture of LLM-generated text detection task. This task is a binary classification task that detects whether the provided text is generated by LLMs or written by humans.
}
\label{fig:detection-task}
\end{figure*}

Detecting LLM-generated text is an intricate challenge. Generally speaking, humans struggle to discern between LLM-generated text and human-written text~\cite{uchendu2021turingbench,dou2021gpt,clark2021all,DBLP:journals/corr/abs-2303-17650,Soni2023ComparingAS}, and their capability to distinguish such texts exceeds random classification only slightly. \autoref{tab:examples} offers some examples where LLM-generated text often is extremely close to human-written text and can be difficult to distinguish. When LLMs generate fabricated details, discerning their origins and veracity remains equally challenging. 

\renewcommand\arraystretch{1.8}
\begin{table}[!ht]
\caption{Examples of human-written text and LLM-generated text. Text generated by LLMs during normal operation and instances in which they fabricate facts often exhibit no intuitively discernible differences. When LLMs either abstain from providing an answer or craft neutral responses, certain indicators, such as the explicit statement ``I am an AI language model'', may facilitate human adjudication, but such examples are less.}
\fontsize{8}{11}
\resizebox{\textwidth}{!}{
\begin{tabular}{cccc}
\toprule
Type &
  Question &
  Human-written &
  LLMs-generated \\ 
\midrule
\multicolumn{1}{c}{Normal} &
  \begin{tabular}[c]{m{3cm}}Explain what is NLP?\end{tabular} &
  \begin{tabular}[c]{ p{6cm} }Natural language processing (NLP) is an interdisciplinary subfield of linguistics, computer science, and artificial intelligence …\end{tabular} &
  \begin{tabular}[c]{ p{6cm} }Natural language processing (NLP) is a field of computer science, artificial intelligence, and  linguistics that focuses on …\end{tabular} \\ 
Refusal &
  \begin{tabular}[c]{m{3cm}}How is today special?\end{tabular} &
  \begin{tabular}[c]{ p{6cm} }Today's Special is a Canadian children's television show produced by Clive VanderBurgh at TVOntario from 1981 to 1987.\end{tabular} &
  \begin{tabular}[c]{ p{6cm} }I'm sorry, but I am an AI language model and do not have access to current dates or events. Is there anything else I can help you with …\end{tabular} \\ 
Fabricated &
  \begin{tabular}[c]{ p{3cm} }Explain what is NLP based on one publication in the recent literature.\end{tabular} &
  \begin{tabular}[c]{ p{6cm} }In “Natural language processing: state of the art, current trends and challenges”, NLP is summarized as a discipline that uses various algorithms, tools and methods to …\end{tabular} &
  \begin{tabular}[c]{ p{6cm} }NLP is a multidisciplinary field at the intersection of computer science, linguistics, and ai, as described in a recent peer-reviewed publication titled "Natural Language Processing: A Comprehensive Overview and Recent Advances" (2023) ...\end{tabular} \\ 
\bottomrule
\end{tabular}
}
\label{tab:examples}
\end{table}

Recent studies \cite{guo2023close, ma2023abstract, munoz2023contrasting, giorgi2023slept,DBLP:journals/corr/abs-2306-04537} have highlighted significant disparities between human-written and LLM-generated text, such as ChatGPT. \textcolor{black}{The differences between LLM-generated text and human-written text are not merely within the scope of individual word choice \cite{DBLP:journals/corr/abs-2306-04537}, but also manifest in stylistic dimensions, such as syntactical simplicity, use of passive voice, and narrativity.} Notably, LLM-generated text often exhibits qualities of enhanced organization, logical structure, formality, and objectivity in comparison to human-written text. Additionally, LLMs frequently produce extensive and comprehensive responses, characterized by a lower prevalence of bias and harmful content. Nevertheless, they occasionally introduce nonsensical or fabricated details.
Linguistically, LLM-generated text tends to be about twice the length of human-written text but exhibits a more limited vocabulary. 
LLMs exhibit a higher frequency of noun, verb, determiner, adjective, auxiliary, coordinating conjunction, and particle word categories compared to humans, and less adverb and punctuation, incorporating more deterministic, conjunctive, and auxiliary structures in their syntax. Additionally, LLM-generated text often conveys less emotional intensity and exhibits clearer presentation than human writings, a phenomenon possibly linked to an inherent positive bias in LLMs \cite{giorgi2021characterizing,markowitz2023linguistic,DBLP:journals/corr/abs-2301-13852}. Although there are slightly different statistical gaps on different datasets, it is clear that the difference between LLM-generated text and human-written text clearly exists, because the statistical results of the difference in language features and human visual perception are consistent. \textcolor{black}{\citet{DBLP:journals/corr/abs-2304-04736} have further substantiated the view by reporting on the detectability of text generated by LLMs, including the high-performance models such as GPT-3.5-Turbo and GPT-4 \cite{DBLP:journals/corr/abs-2309-08913}, while \citet{DBLP:conf/emnlp/ChakrabortyTZGK23} introduced an AI Detectability Index to further rank models according to their detectability.}

In this survey, we begin by providing definitions for human-written text, LLM-generated text, and the detection task.

\paragraph{Human-written Text} is characterized as the text crafted by individuals to express thoughts, emotions, and viewpoints. This encompasses articles, poems, and reviews, among others, typically reflecting personal knowledge, cultural milieu, and emotional disposition, spanning the entirety of the human experience.

\paragraph{LLM-generated Text} is defined as cohesive, grammatically sound, and pertinent content generated by LLMs. These models are trained extensively on NLP techniques, utilizing large datasets and machine learning methodologies.
The quality and fidelity of the generated text typically depend on the scale of the model and the diversity of training data.

\paragraph{LLM-generated Text Detection Task} is conceptualized as a binary classification task, aiming to ascertain if a given text is generated by an LLM. The formal representation of this task is given by the subsequent equation:

\begin{equation}
D(x) =\left\{
\begin{array}{rcl}
1  & & \text{if } x \text{ generated by LLMs} \\
0  & & \text{if } x \text{ written by human}
\end{array} \right.
\end{equation}
\noindent where $D(x)$ represents the detector, and $x$ is the text to be detected.

\subsection{LLMs Text Generation and Confusion Sources} 
\subsubsection{Generation Mechanisms of LLMs}

The text generation mechanism of LLMs operates by sequentially predicting subsequent tokens. Rather than producing an entire paragraph instantaneously, LLMs methodically construct text one word at a time. Specifically, LLMs decode subsequent tokens in a textual sequence, taking into account both the input sequence and previously decoded tokens. Assume that the total time step is $T$, the current time step is $t$, the input text or tokenised sequence is: $ X_T = \{x_1, x_2, ... x_T\} $, and the previous output sequence is $Y_{t-1} = \{y_1, y_2, ... y_{t-1}\}$. At this point, the output next word $y_t$ can be expressed as:

 \begin{equation}
     y_t \sim P(y_t | Y_{t-1},X_T) = softmax(w_o \cdot h_t)
 \end{equation}
\noindent $h_t$ is the hidden state of the model at time step $t$, $w_o$ is the output matrix, the softmax function is used to obtain the probability distribution of the vocabulary, and $y_t$ is sampled from the probability distribution of the vocabulary $P(y_t | Y_{t-1}, X_T)$. The joint probability function for the final output sequence can be modeled and represented as:

 \begin{equation}
     Y_T = \{y_1, y_2, ... , y_T\}
 \end{equation}

The quality of the decoded text is intrinsically tied to the chosen decoding strategy. Given that the model constructs text sequentially, the quality of generated text hinges on the method used to select the subsequent word from the vocabulary, that is, how $y_t$ is sampled from the probability distribution of vocabulary. The predominant decoding techniques encompass greedy search, beam search, top-k sampling, and top-p sampling. \autoref{tab:decoding-methods} offers a comparison of the underlying principles, along with the respective merits and drawbacks, of these decoding methods. This comparison aids in elucidating the text generation process of LLMs and the specific characteristics of the text they produce.

\begin{table}[!ht]
\caption{The core ideas of different text decoding strategies, as well as their advantages and disadvantages. Greedy Search uses a simple greedy strategy, considering only the current highest probability token at each step, which is simple and fast but lacks diversity. Beam Search allows for multiple candidates to be considered, which improves the quality of the text but tends to generate duplicates. Top-K Sampling increases diversity but has difficulty controlling the quality of generation. Top-P Sampling relies on the shape of the probability distribution to determine the set of tokens to sample, which is coherent, but diversity is correlated with the parameter $P$.}
\renewcommand\arraystretch{1.2}
\fontsize{8}{11}\selectfont
\centering
\resizebox{0.98\textwidth}{!}{
\begin{tabular}{p{2.1cm} p{4.0cm} p{3cm} p{4cm}}
\toprule
Strategy &
Core Idea &
Advantages &
Drawbacks \\
\midrule
Greedy Search &
Only the token with the highest current probability is considered at each step.  &
Fast and simple.  &
Easy to fall into local optimality, lack of diversity, unable to deal with uncertainty. 
\\
Beam Search \cite{graves2012sequence} &
Several more candidates can be considered at each step. &
Improvement of text quality and flexibility. &
Tend to generate repetitive fragments, work poorly in open generation domains, unable to handle uncertainty. 
\\
Top-K Sampling \cite{fan2018hierarchical} &
Samples among the K most likely words at each step. &
Increase diversity and be able to deal with uncertainty. &
Difficulty in controlling the quality of generation, which may result in incoherent text. \\
Top-P Sampling \cite{holtzman2019curious} &
Use the shape of the probability distribution to determine the set of tokens to be sampled from &
Coherence and the ability to deal with uncertainty. &
Dependent on the quality of the model predictions, diversity is related to the parameter $P$.\\
\bottomrule
\end{tabular}
}
\label{tab:decoding-methods}
\end{table}

\subsubsection{Sources of LLMs' Strong Generation Capabilities}

The burgeoning growth in model size, data volume, and computational capacity has significantly enhanced the capabilities of LLMs. Beyond a specific model size, certain abilities manifest that are not predictable by scaling laws. These abilities, absent in smaller models but evident in LLMs, are termed ``Emergent Abilities'' of LLMs.

\paragraph{In-Context Learning (ICL)} The origins of ICL capabilities remain a topic of ongoing debate \cite{dai2023can}. However, this capability introduces a paradigm where model parameters remain unchanged, and only the design of the prompt is modified to elicit desired outputs from LLMs. This concept was first introduced in the GPT-3 \cite{NEURIPS2020_1457c0d6}. \citet{NEURIPS2020_1457c0d6} posited that the presence of ICL is fundamental for the swift adaptability of LLMs across a diverse set of tasks. With just a few examples, LLMs can effectively tackle downstream tasks, obviating the previous BERT model-based approach that relied on pretraining followed by fine-tuning for specific tasks \cite{raffel2020exploring}.

\paragraph{Alignment of Human Preference} Although LLMs can be guided to generate content using carefully designed prompts, the resulting text might lack control, potentially leading to the creation of misleading or biased content \cite{zhang2023siren}. The primary concern of these models lies in predicting subsequent words to form coherent sentences based on vast corpora, rather than ensuring that the content generated is both beneficial and innocuous to humans. To address these shortcomings, OpenAI introduced the Reinforcement Learning from Human Feedback (RLHF) approach, as detailed in \cite{ouyang2022training} and \cite{lambert2022illustrating}. This approach begins by fine-tuning LLMs using data from user-directed quizzes and subsequently evaluating the model's outputs with human assessors. Simultaneously, a reward function is established, and the LLM is further refined using the Proximal Policy Optimization (PPO) algorithm~\cite{DBLP:journals/corr/SchulmanWDRK17}. The end result is a model that aligns with human values, understands human instructions, and genuinely assists users.

\paragraph{Complex Reasoning Capabilities} While LLMs' ICL and alignment capability facilitate meaningful interactions and assistance, their performance tends to degrade in tasks demanding logical reasoning and heightened complexity. \citet{wei2022chain} observed that encouraging LLMs to produce more intermediate steps through a Chain of Thought (CoT) can enhance their effectiveness. Tree of Thoughts (ToT) \cite{yao2023tree} and Graph of Thoughts (GoT) \cite{besta2023graph} are extensions of this methodology. Both strategies augment LLM performance on intricate tasks by amplifying the computational effort required for the model to deduce an answer.

\subsection{Why Do We Need to Detect Text Generated by LLMs?}
As LLMs undergo iterative refinements and reinforcement learning through human feedback, their outputs become increasingly harmonized with human values and preferences. This alignment facilitates broader acceptance and integration of LLM-generated text into everyday life. 
The emergence of various AI tools has played a significant role in fostering intuitive human-AI interactions and democratizing access to the advanced capabilities of previously esoteric models.
From interactive web-based assistants like ChatGPT,\footnote{\url{https://chat.openai.com/}} to search engines enhanced with LLM technology like the contemporary version of Bing,\footnote{\url{https://www.bing.com/}} to specialized tools like Coplit,\footnote{\url{https://github.com/features/copilot/}} and Scispeace\footnote{\url{https://typeset.io/}}  that assist professionals in code generation and scientific research, LLMs have subtly woven into the digital fabric of our lives, propagating their content across diverse platforms.

However, it is important to acknowledge that for the majority of users, LLMs and their applications are still considered black-box AI systems. For individual users, this often serves as a benign efficiency boost, sidestepping laborious retrieval, and summarization. However, within specific contexts and against the backdrop of the broader digital landscape, it becomes crucial to discern, filter, or even exclude LLM-generated text. It is important to emphasize that not all situations call for the detection of LLM-generated text. Unnecessary detection can lead to consequences such as system inefficiencies and inflated development costs. Generally, detecting LLM-generated text might be superfluous when:
\begin{itemize}
\item The utilization of LLMs poses minimal risk, especially when they handle routine, replicable tasks.
\item The spread of LLM-generated text is confined to predictable, limited domains, like closed information circles with few participants.
\end{itemize}

\begin{figure*}[ht]
\centering
\includegraphics[width=0.65\textwidth, trim=0 0 0 0]{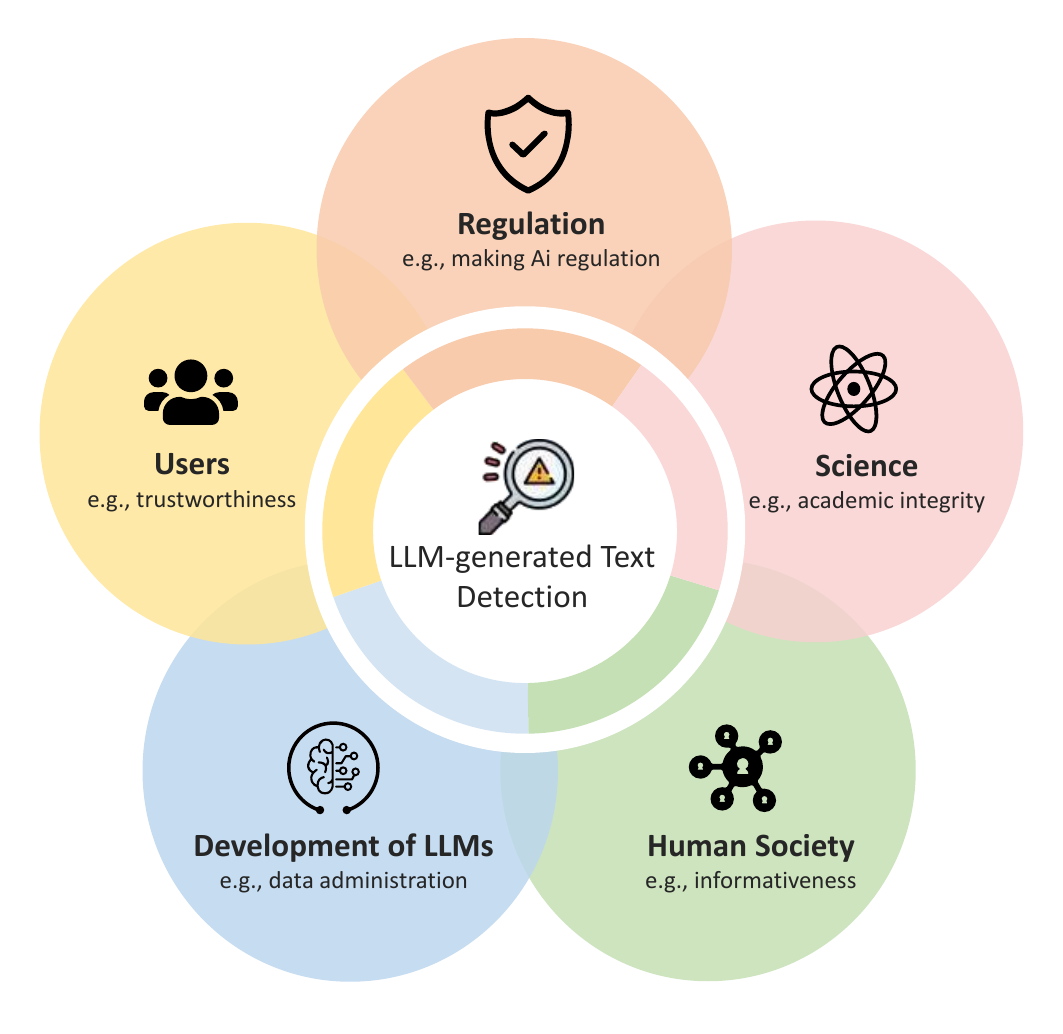}
\caption{The most critical reasons why LLM-generated text detection is needed urgently. We discussed it from five perspectives: Regulation, Users, Developments, Science, and Human Society.
}
\label{fig:perspect}
\end{figure*}

Drawing upon the literature reviewed in this study, the rationale behind detecting LLM-generated text can be elucidated from multiple vantage points, as illustrated in \autoref{fig:perspect}. The delineated perspectives are, in part, informed by the insights presented in \cite{gade2020explainable} and \cite{saeed2023explainable}.

While the list provided above may not be exhaustive and some facets may intersect or further delineate as LLMs and AI systems mature, we posit that these points underscore the paramount reasons for the necessity of detecting text generated by LLMs.

\paragraph{Regulation} As AI tools, often characterized as black boxs, the inclusion of LLM-generated text in creative endeavors raises significant legal issues. A pressing concern is the eligibility of LLM-generated texts for intellectual property rights protection, a subject still mired in debate \cite{epstein2023art, wikipedia-llms-copyright}, although the \emph{EU AI Act}\footnote{\url{https://artificialintelligenceact.eu/the-act/}} has begun to continuously improve to regulate the use of AI. The main challenges arise from issues such as ownership of the training data used by the AI in generating output and how to determine how much human involvement is enough to make the work theirs. The prerequisite for copyright protection for AI supervision and AI-generated content is that human creativity in the materials used to train AI systems can be distinguished, so as to further promote the implementation of more complete legal supervision.

\paragraph{Users} LLM-generated text, refined through various alignment methods, is progressively aligning with human preferences. This content permeates numerous user-accessible platforms, including blogs and Questions \& Answers (Q\&A) forums. However, excessive reliance on such content can undermine user trust in AI systems and, by extension, digital content as a whole. In this context, the role of LLM-generated text detection becomes crucial as a gatekeeper to regulate the prevalence of LLM-generated text online.

\paragraph{Developments} With the evolving prowess of LLMs, \citet{li2023self} suggested that LLMs can self-assess and even benchmark their own performances. Due to its excellent text generation performance, LLMs are also used to construct many training data sets through preset instructions \cite{alpaca}. However, \citet{alemohammad2023selfconsuming} posited that this ``Self-Consuming'' paradigm may engender a homogenization in LLM-generated texts, potentially culminating in what is termed as ``LLM Autophagy Disorder'' (MAD). If LLMs heavily rely on web-sourced data for training, and a significant portion of this data originates from LLM outputs, it could hinder their long-term progress.

\paragraph{Science} The relentless march of human progress owes much to the spirit of scientific exploration and discovery. However, the increasing presence of LLM-generated text in academic writing \cite{majovsky2023artificial} and the use of LLM-originated designs in research endeavors raise concerns about potentially diluting human ingenuity and exploratory drive. \textcolor{black}{At the same time, it could also undermine the ability of higher education to validate student knowledge and comprehension, and diminish the academic reputation of specific higher education institutions \cite{DBLP:journals/corr/abs-2305-13934}.} Although current methodologies may have limitations, further enhancements in detection capabilities will strengthen academic integrity and preserve human independent thinking in scientific research. 

\paragraph{Human Society} From a societal perspective, analyzing the implications of LLM-generated text reveals that these models essentially mimic specific textual patterns while predicting subsequent tokens. If used improperly, these models have the potential to diminish linguistic diversity and contribute to the formation of information silos within societal discourse. In the long run, detecting and filtering LLM-generated text is crucial for preserving the richness and diversity of human communication, both linguistically and informatively.
\section{Related Works and Our Investigation}
\subsection{Related Works}
\label{related_works}

\textcolor{black}{
 The comprehensive review article by \citet{beresneva2016computer} represents the first extensive survey of methods for detecting computer-generated text. At that time, the detection process was relatively simple, mainly focusing on machine translation text detection and employing simple statistical methods for detection. The emergence of autoregressive models has significantly increased the complexity of text detection tasks. \citet{jawahar2020automatic} provide a detailed survey on the detection of machine-generated text, establishing a foundational context for the field with an emphasis on the SOTA generative models prevalent at the time, such as GPT-2. The subsequent release of ChatGPT sparked a surge of interest in LLMs, and signified a major shift in research directions. In response to the rapid challenges posed by LLM-generated text, the NLP community has recently embarked on extensive research to devise robust detection mechanisms and explore the dynamics of detector evasion techniques, aiming to seek effective solutions. The recent survey by \citet{DBLP:journals/access/CrothersJV23,DBLP:journals/corr/abs-2309-07689} provide new reviews of LLM-generated text detection, but we contend that these reviews are not advanced enough and the summary of detection methods needs to be improved. \citet{tang2023science} present another survey, categorizing detection methods into black-box detection and white-box detection, and highlighting cutting-edge technologies such as watermarking, but the review could benefit from a more comprehensive analysis and critical evaluation.  \citet{ghosal2023survey} discuss the current attacks and defenses of AI-generated text detectors and provide a thorough inductive analysis. Nevertheless, the discussion could be enriched with a more detailed examination of task motivation, data resources, and evaluation methodologies.
}

\textcolor{black}{In this article, we strive to provide a more comprehensive and insightful review of the latest research on LLM-generated text detection, enriched with thoughtful analysis. We highlight the strengths of our review in comparison to others:}

\textcolor{black}{\begin{itemize}
    \item Systematic and Comprehensive Review: Our survey offers an extensive exploration of LLM-generated text detection, covering the task's description and underlying motivation, benchmarks and datasets, various detection and attack methods, evaluation frameworks, the most pressing challenges faced today, potential future directions, and a critical examination of each aspect.
    \item In-depth Analysis of Detection Mechanisms: We provide a detailed overview of detection strategies, from traditional approaches to the latest research, and systematically evaluate their effectiveness, strengths, and weaknesses in the current environment of LLMs.
    \item More Pragmatic Insights. Our discussion delves into research questions with practical implications, such as how model size affects detection capabilities, the challenges of identifying text that is not purely generated by LLMs, and the lack of effective evaluation framework.
\end{itemize}}

\textcolor{black}{
In summary, we firmly believe that this review is more systematic and comprehensive than existing works. More importantly, our critical discussion not only provides guidance to new researchers but also imparts valuable insights into established works within the field.
}

\subsection{Systematic Investigation and Implementation}

\begin{table}[!ht]
\caption{Overview of the diverse databases and search engines utilized in our research, along with the incorporated search schemes and the consequent results obtained. Google Scholar predominates as the search engine yielding the maximum number of retrievable documents. Upon meticulous examination, it is observed that a substantial portion of the documents originate from ArXiv, primarily shared by researchers.}
\fontsize{8}{11}\selectfont
\begin{tabular}{ p{2.6cm} p{4cm} p{4cm} p{1cm}}
\toprule
\textbf{Databases} & \textbf{Search Engine} & \textbf{Search Scheme} & \textbf{Retrieved} \\
\midrule
Google Scholar & \href{https://scholar.google.com/}{https://scholar.google.com/} & Full Text & 210 \\

ArXiv & \href{https://arxiv.org/}{https://arxiv.org/} & Full Text & N/A$^{\rm a}$ \\

Scopus & \href{https://www.scopus.com/}{https://www.scopus.com/} & TITLE-ABS-KEY: ( Title, Abstract, Author Keywords, Indexed Keywords ) & 133 \\

Web of Science & \href{https://www.webofscience.com/}{https://www.webofscience.com/} & Topic: ( Searches Title, Abstract, Author Keywords, Keywords Plus ) & 92 \\

IEEE Xplore & \href{https://ieeexplore.ieee.org/}{https://ieeexplore.ieee.org/} & Full Text & 49 \\

Springer Link & \href{https://link.springer.com/}{https://link.springer.com/} & Full Text & N/A$^{\rm a}$ \\

ACL Anthology & \href{https://aclanthology.org/}{https://aclanthology.org/} & Full Text & N/A$^{\rm a}$ \\

ACM Digital Library & \href{https://dl.acm.org/}{https://dl.acm.org/} & Title & N/A$^{\rm b}$ \\
\bottomrule
\end{tabular}
\tabnote{$^{\rm a}$ Search engines cannot use all keywords in a single search string. Therefore the retrieved results are inaccurate and there may be duplicate results of thesis queries.}
\tabnote{$^{\rm b}$ The search engine retrieved an inaccurate number of papers that were weakly related to our topic.}
\label{tab:search-engines}
\end{table}

Our survey employed the System for Literature Review (SLR) as delineated by \citet{keele2007guidelines}, a methodological framework designed for evaluating the extent and quality of extant evidence pertaining to a specified research question or topic. Offering a more expansive and accurate insight compared to conventional literature reviews, this approach has been prominently utilized in numerous scholarly surveys, as evidenced by \citet{murtaza2020deep,saeed2023explainable}. The research questions guiding our SLR were as follows: \\

{\emph{What} are the prevailing methods for detecting LLM-generated text, and \emph{what} are the main challenges associated with these methods?}

\begin{figure}
\centering
\pgfplotsset{compat=1.16}
\definecolor{colorA}{RGB}{230, 230, 250}
\definecolor{colorB}{RGB}{255, 218, 185}
\begin{tikzpicture}
\begin{axis}[
    ybar,                   
    enlarge x limits=0.15,    
    ylabel={Number of publications},       
    xlabel={Year},   
    symbolic x coords={2019, 2020, 2021, 2022, 2023}, 
    xtick=data,       
    bar width=15pt,
    ymin=0,
    legend style={at={(0.5,-0.15)},anchor=north,legend columns=-1},
    ylabel near ticks,
    ]
\addplot[color=colorA,fill=colorA] coordinates {(2019,4) (2020,7) (2021,3) (2022,10) (2023,58)};
\addplot[mark=*, color=colorB,smooth,line width=2pt] coordinates {(2019,4) (2020,8) (2021,3) (2022,10) (2023,58)};
\end{axis}
\end{tikzpicture}
\caption{The distribution by year of the last 5 years of literature obtained from the screening is plotted. The number of published articles obtain significant attention in 2023.
}
\label{fig:distribution}
\end{figure}

Upon delineating the research problems, our study utilized search terms directly related to the research issue, specifically: ``LLM-generated text detection'', ``machine-generated text detection'', ``AI-written text detection'', \textcolor{black}{``authorship attribution''}, and ``deepfake text detection''. These terms were strategically combined using the Boolean operator OR to formulate the following search string: (``LLM-generated text detection'' OR ``machine-generated text detection'' OR ``AI-written text detection'' OR \textcolor{black}{``authorship attribution''} OR ``deepfake text detection''). Subsequently, employing this search string, we engaged in a preliminary search through pertinent and authoritative electronic dissertation databases and search engines. Our investigation mainly focused on scholarly articles that were publicly accessible prior to November 2023. \autoref{tab:search-engines} outlines the sources used and provides an overview of our results.

Subsequently, we established the ensuing criteria to scrutinize the amassed articles:

\begin{itemize}
\item The article should be a review focusing on the methods and challenges pertinent to LLM-generated (machine-generated/AI-written) text detection.
\item The article should propose a methodology specifically designed for the detection of LLM-generated (machine-generated/AI-written) text.
\item The article should delineate challenges and prospective directions for future research in the domain of text generation for LLMs.
\item The article should articulate the necessity and applications of LLM-generated text detection.
\end{itemize}

If any one of the aforementioned four criteria was met, the respective work was deemed valuable for our study. Following a process of de-duplication and manual screening, we identified 83 pertinent pieces of literature. The distribution trend of these works, delineated by year, is illustrated in \autoref{fig:distribution}. Notably, the majority of relevant research on LLM-generated text detection was published in the year 2023 (as shown in \autoref{fig:distribution}), underscoring the vibrant development within this field and highlighting the significance of our study. In subsequent sections, we provide a synthesis and analysis of the data (see \autoref{data}), primary detectors (see \autoref{detectors}), evaluation metrics (see \autoref{metrics}), issues (see \autoref{issues}), and future research directions (see \autoref{future_research_directions}) pertinent to LLM-generated text detection. The comprehensive structure of the survey is outlined in \autoref{tab:structure}, offering a detailed overview of the organization of our review.

\begin{table}[!ht]
\caption{Summary of content organisation of this survey.}
\renewcommand\arraystretch{1.2}
\fontsize{8}{11}\selectfont
\centering
\resizebox{1\textwidth}{!}{
\begin{tabular}{ p{1.2cm} p{2.5cm} p{8.0cm}}
\toprule
Section &
  Topic &
  Content \\
\midrule
Section 4 &
  Data &
  Datasets and Benchmarks for LLM-generated Text Detection, Other Datasets Easily Extended to Detection Tasks and Challenge of datsets for LLM-generated Text Detection. \\
Section 5 &
  Detectors &
  Watermarking Technology, Statistics-Based Detectors, Neural-Based Detectors, and Human-Assisted Methods  \\
Section 6 &
  Evaluation Metrics &
  Accuracy, Precision, Recall, False Positive Rate, True Negative Rate, False Negative Rate, F1-Score, and Area under the ROC curve (AUROC). \\
Section 7 &
  Issues &
  Out of Distribution Challenges, Potential Attacks,  Real-world Data Issues, Impact of Model Size on Detectors, and  Lack of Effective Evaluation Framework \\
Section 8 &
  Future Directions &
  Building Robust Detectors with Attacks, Enhancing the Efficacy of Zero-Shot Detectors, Optimizing Detectors for Low-resource Environments, Detection for Not purely LLM-generated Text, Constructing Detectors Amidst Data Ambiguity, Developing Effective Evaluation Framework Aligned With Real-World, and Constructing Detectors with  Misinformatio Discrimination Capabilities. \\
Section 9 &
  Conclusion &
  -\\
\bottomrule
\end{tabular}
}
\label{tab:structure}
\end{table}
\section{Data}
\label{data}
High-quality datasets are pivotal for advancing research in the LLM-generated text detection task. These datasets enable researchers to swiftly develop and calibrate efficient detectors and establish standardized metrics for evaluating the efficacy of their methodologies. However, procuring such high-quality labeled data often demands substantial financial, material, and human resources. Presently, the development of datasets focused on detecting LLM-generated text is in its nascent stages, plagued by issues such as limited data volume and sample complexity, both crucial for crafting robust detectors. In this section, we first introduce popular datasets employed for training LLM-generated text detectors. Additionally, we highlight datasets from unrelated domains or tasks, which, though not initially designed for detection tasks, can be repurposed for various detection scenarios, which is a prevailing strategy in many contemporary detection studies. \textcolor{black}{We subsequently introduce benchmarks for verifying the effectiveness of LLM-generated text detectors, which are carefully designed to evaluate the performance of the detector from different perspectives.} Lastly, we evaluate these training datasets and benchmarks, identifying current shortcomings and challenges in dataset construction for LLM-generated text detection, aiming to inform the design of future data resources.

\subsection{Training}
\subsubsection{Detection Datasets}

\textcolor{black}{Massive and high-quality datasets can assist researchers in rapidly training their detectors. We thoroughly organize and compare datasets that are widely used and have potential, refer to \autoref{tab:detection_datasets}. Given that different studies focus on various practical issues, our aim is to facilitate researchers in conveniently selecting high-quality datasets that meet their specific needs through our comprehensive review work.}

\begin{table}[!ht]
\caption{Summary of Detection Datasets for LLM-generated text detection.}
\renewcommand\arraystretch{1.2}
\fontsize{8}{11}\selectfont
\centering
\resizebox{1\textwidth}{!}{
\begin{tabular}{ p{3.0cm} p{0.7cm} p{0.6cm} p{0.6cm} p{4.0cm} p{1.5cm} p{1.3cm} p{4cm} }
\toprule
Corpus & Use & Human & LLMs & LLMs Type & Language & Attack & Domain \\
\midrule
HC3 \cite{guo2023close} & train & \textasciitilde80k & \textasciitilde43k & ChatGPT & English, Chinese & - & Web Text, QA, Social Media \\
CHEAT \cite{yu2023cheat} & train & \textasciitilde15k & \textasciitilde35k & ChatGPT & English & Paraphrase & Scientific Writing \\
HC3 Plus \cite{DBLP:journals/corr/abs-2309-02731} & train valid test & \multicolumn{2}{c}{\parbox[t]{0.6cm}{\textasciitilde95k \textasciitilde10k \textasciitilde38k}} & GPT-3.5-Turbo & Englilsh, Chinese & Paraphrase & News Writing, Social Media \\
OpenLLMText \cite{DBLP:conf/emnlp/ChenKZLSR23} & train, valid, test & \textasciitilde52k \textasciitilde8k \textasciitilde8k
& \textasciitilde209k \textasciitilde33k \textasciitilde33k & ChatGPT, PaLM, LLaMA, GPT2-XL & English & - & Web Text \\

GROVER Dataset \cite{zellers2019defending} & train & \multicolumn{2}{c}{\textasciitilde24k} & Grover-Mega & English & - & News Writing \\
TweepFake \cite{Fagni_2021} & train & \textasciitilde12k & \textasciitilde12k & GPT-2, RNN, Markov, LSTM, CharRNN & English & - & Social Media \\
GPT-2 Output Dataset\footnote{\url{https://github.com/openai/gpt-2-output-dataset}} & train test & \textasciitilde250k \textasciitilde5k& \textasciitilde2000k \textasciitilde40k& GPT-2 (small, medium, large, xl) & English & - & Web Text\\
ArguGPT \cite{liu2023argugpt} & train valid test & \multicolumn{2}{c}{\parbox[t]{0.6cm}{\textasciitilde6k 700 700}} & GPT2-Xl, Text-Babbage-001, Text-Curie-001, Text-Davinci-001, Text-Davinci-002, Text-Davinci-003, GPT-3.5-Turbo & English & - & Scientific writing \\
DeepfakeTextDetect \cite{li2023deepfake} & train valid test &  \multicolumn{2}{c}{\parbox[t]{0.6cm}{\textasciitilde236k \textasciitilde56k \textasciitilde56k}} & GPT (Text-Davinci-002, Text-Davinci-003, GPT-Turbo-3.5),
LLaMA (6B, 13B, 30B, 65B), GLM-130B,
FLAN-T5 (small, base, large, xl, xxl),
OPT(125M, 350M, 1.3B, 2.7B, 6.7B, 13B, 30B, iml1.3B, iml-30B), T0 (3B,
11B), BLOOM-7B1, GPT-J-6B, GPT-NeoX-20B)& English & Paraphrase & Social Media, News Writing, QA, Story Generation, Comprehension and Reasoning, Scientific writing \\
\bottomrule
\end{tabular}
}
\label{tab:detection_datasets}
\end{table}

\paragraph{HC3}
The Human ChatGPT Comparison Corpus (HC3) \cite{guo2023close} stands as one of the initial open-source efforts to compare ChatGPT-generated text with human-written text. It involves collecting both human and ChatGPT responses to identical questions. Due to its pioneering contributions in this field, the HC3 corpus has been utilized in numerous subsequent studies as a valuable resource. The corpus offers datasets in both English and Chinese. Specifically, HC3-en comprises 58k human responses and 26k ChatGPT responses, derived from 24k questions, which primarily originate from the ELI5 dataset, WikiQA dataset, Crawled Wikipedia, Medical Dialog dataset, and FiQA dataset. On the other hand, HC3-zh encompasses a broader spectrum of domains, featuring 22k human answers and 17k ChatGPT responses. The data within HC3-zh spans seven sources: WebTextQA, BaikeQA, Crawled BaiduBaike, NLPCC-DBQA dataset, Medical Dialog dataset, Baidu AI Studio, and the LegalQA dataset. However, it is pertinent to note some limitations of the HC3 dataset, such as the lack of diversity in prompts used for data creation.

\paragraph{CHEAT} The CHEAT dataset \cite{yu2023cheat} stands as the most extensive publicly accessible resource dedicated to detecting spurious academic content generated by ChatGPT. It includes human-written academic abstracts sourced from IEEE Xplore, with an average abstract length of 163.9 words and a vocabulary size of 130k words. Following the ChatGPT generation process, the dataset contains 15k human-written abstracts and 35k ChatGPT-generated summaries. To better simulate real-world applications, the outputs were guided by ChatGPT for further refinement and amalgamation. The ``polishing'' process aims to simulate users who may seek to bypass plagiarism detection by improving the text, while ``blending'' represents scenarios where users might combine manually drafted abstracts with those seamlessly crafted by ChatGPT to elude detection mechanisms. Nevertheless, a limitation of the CHEAT dataset is its focus on narrow academic disciplines, overlooking cross-domain challenges, which stems from constraints related to its primary data source.

\textcolor{black}{
\paragraph{HC3 Plus} 
HC3 Plus \cite{DBLP:journals/corr/abs-2309-02731} represents an enhanced version of the original HC3 dataset, introducing an augmented section named \textit{HC3-SI}. This new section specifically targets tasks requiring semantic invariance, such as summarization, translation, and paraphrasing, thus extending the scope of HC3. To compile the human-written text corpus for HC3-SI, data was curated from several sources, including the CNN/DailyMail dataset, Xsum, LCSTS, the CLUE benchmark, and datasets from the Workshop on Machine Translation (WMT). Simultaneously, the LLM-generated texts were generated using GPT-3.5-Turbo. The expanded English dataset now includes a training set of 95k samples, a validation set of 10k samples, and a test set of 38k samples. The Chinese dataset, in comparison, contains 42k training samples, 4k for validation, and 22k for testing. Despite these expansions, HC3-SI still mirrors HC3's approach to data construction, which is somewhat monolithic and lacks diversity, particularly in the variety of LLMs and the use of complex and varied prompts for generating data.
}

\paragraph{OpenLLMText}
\textcolor{black}{
The OpenLLMText dataset \cite{DBLP:conf/emnlp/ChenKZLSR23} is derived from four types of LLMs: GPT-3.5, PaLM, LLaMA-7B, and GPT2-1B (also known as GPT-2 Extra Large). The samples from GPT2-1B are sourced from the GPT-2 Output dataset, which OpenAI has made publicly available. Text generation from GPT-3.5 and PaLM was conducted using the prompt ``Rephrase the following paragraph paragraph by paragraph: [Human\_Sample],'' while LLaMA-7B generated text by completing the first 75 tokens of human samples. The dataset comprises a total of 344k samples, including 68k written by humans. It is divided into training, validation, and test sets at 76\%, 12\%, and 12\%, respectively. Notably, this dataset features LLMs like PaLM, which are commonly used in everyday applications. However, it does not fully capture the nuances of cross-domain and multilingual text, which limits its usefulness for related research.}

\paragraph{TweepFake Dataset} TweepFake \cite{Fagni_2021} is a foundational dataset designed for the analysis of fake tweets on Twitter, derived from both genuine and counterfeit accounts. It encompasses a total of 25k tweets, with an equal distribution between human-written and machine-generated samples. The machine-generated tweets were crafted using various techniques, including GPT-2, RNN, Markov, LSTM, and CharRNN. While TweepFake continues to be a dataset of choice for many scholars, those working with LLMs should critically assess its relevance and rigor in light of evolving technological capabilities.

\paragraph{GPT2-Output Dataset} The GPT2-Output Dataset,\footnote{\url{https://github.com/openai/gpt-2-output-dataset}} introduced by OpenAI, is based on 250k documents sourced from the WebText test set for its human-written text. Regarding the LLM-generated text, the dataset includes 250k randomly generated samples using a temperature setting of 1 without truncation and an additional 250k samples produced with Top-K 40 truncation. This dataset was conceived to further research into the detectability of the GPT-2 model. However, a notable limitation lies in the insufficient complexity of the dataset, marked by the uniformity of both the generative models and data distribution.

\paragraph{GROVER Dataset} The GROVER Dataset, introduced by \citet{zellers2019defending}, is styled after news articles. Its human-written text is sourced from RealNews, a comprehensive corpus of news articles derived from Common Crawl. The LLM-generated text is produced by Grover-Mega, a transformer-based news generator with 1.5 billion parameters. A limitation of this dataset, particularly in the current LLM landscape, is the uniformity and singularity of both its generative model and data distribution.

\paragraph{ArguGPT Dataset}
The ArguGPT Dataset \cite{liu2023argugpt} is specifically designed for detecting LLM-generated text in various academic contexts such as classroom exercises, TOEFL, and GRE writing tasks. It comprises 4k argumentative essays, generated by seven distinct GPT models. Its primary aim is to tackle the unique challenges associated with teaching English as a second language.

\paragraph{DeepfakeTextDetect Dataset}
Attention is also drawn to the DeepfakeTextDetect Dataset \cite{li2023deepfake}, a robust platform tailored for deepfake text detection. The dataset combines human-written text from ten diverse datasets, encompassing genres like news articles, stories, scientific writings, and more. It features texts generated by 27 prominent LLMs, sourced from entities such as OpenAI, LLaMA, and EleutherAI. Furthermore, the dataset introduces an augmented challenge with the inclusion of text produced by GPT-4 and paraphrased text.

\subsubsection{Potential Datasets}
Constructing a dataset from scratch that encompasses both human-written and LLM-generated text can indeed be a resource-intensive endeavor. Recognizing the diverse requirements for LLM-generated text detection across scenarios, researchers commonly adapt existing datasets from fields like Q\&A, academic writing, and story generation to represent human-written text. They then produce LLM-generated text for detectors training using methods like prompt engineering and bootstrap complementation. This survey offers a concise classification and overview of these datasets, summarized in \autoref{tab:other-datasets}.

\begin{table*}[!ht]
\caption{Summary of other potential datasets that can easily extended to LLM-generated text detection tasks.}
\centering
\resizebox{\textwidth}{!}{
\begin{tabular}{ l c c c c }
\toprule
\textbf{Corpus} &
  \textbf{Size} &
  \textbf{Source} &
  \textbf{Language} &
  \textbf{Domain} \\
\midrule
XSum \cite{narayan2018don} &
  42k &
  BBC &
  English &
  News Writing \\
SQuAD \cite{rajpurkar2016squad} &
  98.2k &
  Wiki &
  English &
  Question Answering \\
WritingPrompts \cite{fan2018hierarchical} &
  302k &
  Reddit WRITINGPROMPTS &
  English &
  Story Generation \\
Wiki40B \cite{guo2020wiki} &
  17.7m &
  Wiki &
  40+ Languages &
  Web Text \\
PubMedQA \cite{jin2019pubmedqa} &
  211k &
  PubMed &
  English &
  Question Answering \\
Children's Book Corpus \cite{hill2015goldilocks} &
  687k &
  Books &
  English &
  Question Answering \\
Avax Tweets Dataset \cite{muric2105covid} &
  137m &
  Twitter &
  English &
  Social Media \\

Climate Change Dataset \cite{DVN/5QCCUU_2019} &
  4m &
  Twitter &
  English &
  Social Media \\

Yelp Dataset \cite{asghar2016yelp} &
  700k &
  Yelp &
  English &
  Social Media \\

ELI5 \cite{fan2019eli5} &
  556k &
  Reddit &
  English &
  Question Answering \\

ROCStories \cite{mostafazadeh2016corpus} &
  50k &
  Crowdsourcing &
  English &
  Story Generation \\

HellaSwag \cite{zellers2019hellaswag}  &
  70k &
  ActivityNet Captions, Wikihow &
  English &
  Question Answering \\

SciGen \cite{moosavi2021scigen} &
  52k &
  arXiv &
  English &
  Scientific Writing, Question Answering \\

WebText \cite{radford2019language} &
  45m &
  Web &
  English &
  Web Text \\

TruthfulQA \cite{lin2021truthfulqa} &
  817 &
  authors writtEnglish &
  English &
  Question Answering \\

NarrativeQA \cite{kovcisky2018narrativeqa} &
  1.4k &
  Gutenberg3, web &
  English &
  Question Answering \\

TOEFL11 \cite{blanchard2013toefl11} &
  12k &
  TOEFL test &
  11 Languages&
  Scientific writing \\

\multirow{2}{*}{Peer Reviews \cite{kang2018dataset}} & \multirow{2}{*}{14.5k} & NIPS 2013–2017, CoNLL 2016, ACL 2017 &
\multirow{2}{*}{English} &
\multirow{2}{*}{Scientific Writing} \\
 & & ICLR 2017, arXiv 2007–2017 & & \\
\bottomrule
\end{tabular}
}
\label{tab:other-datasets}
\end{table*}

\paragraph{Q\&A}
Q\&A is a prevalent and equitable method for constructing datasets. By posing identical questions to LLMs, one can generate paired sets of human-written and LLM-generated text.

\begin{itemize}
\item \textit{PubMedQA} \cite{jin2019pubmedqa}: This is a biomedical question-and-answer (QA) dataset sourced from PubMed.\footnote{\url{https://www.ncbi.nlm.nih.gov/pubmed/}}

\item \textit{Children Book Corpus} \cite{hill2015goldilocks}: This dataset, derived from freely available books, gauges the capacity of LMs to harness broader linguistic contexts. It challenges models to select the correct answer from ten possible options, based on a context of 20 consecutive sentences. The answer types encompass verbs, pronouns, named entities, and common nouns.

\item \textit{ELI5} \cite{fan2019eli5}: A substantial corpus for long-form Q\&A, ELI5 focuses on tasks demanding detailed responses to open-ended queries. The dataset comprises 270k entries from the Reddit forum ``Explain Like I'm Five'', which offers explanations tailored to the comprehension level of a five-year-old.

\item \textit{TruthfulQA} \cite{lin2021truthfulqa}: This benchmark evaluates the veracity of answers generated by LLMs. It encompasses 817 questions spread across 38 categories such as health, law, finance, and politics. All questions were crafted by humans.

\item \textit{NarrativeQA} \cite{kovcisky2018narrativeqa}: This English-language dataset includes summaries or stories along with related questions aimed at assessing reading comprehension, especially concerning extended documents. Data is sourced from Project Gutenberg \footnote{\url{https://gutenberg.org/}} and web-scraped movie scripts, with hired annotators providing the answers.

\end{itemize}

\paragraph{Scientific Writing}
Scientific writing is frequently explored in real-world research contexts. Given a specific academic topic, LLMs can efficiently generate academic articles or abstracts.
\begin{itemize}
\item \textit{Peer Read} \cite{kang2018dataset}: This represents the inaugural public dataset of scientific peer-reviewed articles, comprising 14.7k draft articles and 10.7k expert-written peer reviews for a subset of these articles. Additionally, it includes the acceptance or rejection decisions from premier conferences such as ACL, NeurIPS, and ICLR.
\item \textit{ArXiv}:\footnote{\url{https://arxiv.org/}} A freely accessible distribution service and repository, ArXiv hosts 2.3 million scholarly articles spanning disciplines like physics, mathematics, computer science, and statistics.
\item \textit{TOEFL11} \cite{blanchard2013toefl11}: A publicly accessible corpus featuring works of non-native English writers from the TOEFL test, it encompasses 1.1k essay samples across 11 languages: Arabic, Chinese, French, German, Hindi, Italian, Japanese, Korean, Spanish, Telugu, and Turkish. These essays are uniformly distributed over eight writing prompts and three score levels (low/medium/high).
\end{itemize}

\paragraph{Story Generation}
LLMs excel in the domain of story generation, with users frequently utilizing story titles and writing prompts to guide the models in their creative endeavors.
\begin{itemize}
\item \textit{WritingPrompts} \cite{fan2018hierarchical}: This dataset comprises 300k human-written stories paired with writing prompts. The data was sourced from Reddit's writing prompts forum, a vibrant online community where members inspire one another by posting story ideas or prompts. Stories in this dataset are restricted in length, ranging between 30 to 1k words, with no words repeated more than 10 times.
\end{itemize}

\paragraph{News Writing}
The task of news article writing can be approached through article summary datasets. LLMs can be instructed either to generate abstracts from the primary text or to generate articles based on provided abstracts. Nonetheless, given the resource constraints, researchers often employ LLMs to generate such datasets by directly reinterpreting or augmenting the existing abstracts and articles.
\begin{itemize}
\item
\textit{Extreme Summarization (XSum)} \cite{narayan2018don}: This dataset contains BBC articles accompanied by concise one-sentence abstracts. It encompasses a total of 225k samples from 2010 to 2017, spanning various domains such as News, Politics, Sports, Weather, Business, Technology, Science, Health, Family, Education, Entertainment, Arts, and more.
\end{itemize}

\paragraph{Web Text}
Web text data primarily originates from platforms like Wikipedia. For web text generation, a common approach is to provide the LLMs with an opening sentence and allow them to continue the narrative. Alternatively, LLMs can be instructed to generate content based on a webpage title.
\begin{itemize}
\item \textit{Wiki-40B} \cite{guo2020wiki}: Initially conceived as a multilingual benchmark for language model training, this dataset comprises text from approximately 19.5 million Wikipedia pages across 40 languages, aggregating to about 40 billion characters. The content has been meticulously cleaned to maintain quality.
\item \textit{WebText} \cite{radford2019language}: Originally utilized to investigate the learning potential of LMs or LLMs, this dataset encompasses 45 million web pages. Prioritizing content quality, the dataset exclusively includes web pages curated or filtered by humans, while deliberately excluding common sources from other datasets, such as Wikipedia.
\end{itemize}

\paragraph{Social Media} Social Media datasets are instrumental in assessing the disparity in subjective expressions between LLM-generated and human-written texts.

\begin{itemize}
\item \textit{Yelp Reviews Dataset} \cite{asghar2016yelp}: Sourced from the 2015 Yelp Dataset Challenge, this dataset was primarily used for classification tasks such as predicting user ratings based on reviews and determining polarity labels. It comprises 1.5 million review text samples.
\item \textit{r/ChangeMyView (CMV) Reddit Subcommunity}:\footnote{\url{https://www.reddit.com/r/changemyview/}} Often referred to as ``Change My View (CMV)'', this subreddit offers a platform for users to debate a spectrum of topics, ranging from politics and media to popular culture, often presenting contrasting viewpoints.
\item \textit{IMDB Dataset}:\footnote{\url{https://huggingface.co/datasets/imdb}} Serving as an expansive film review dataset for binary sentiment classification, it exceeds prior benchmark datasets in volume, encompassing 25k training and 25k test samples.
\item \textit{Avax Tweets Dataset} \cite{muric2105covid}: Designed to examine anti-vaccine misinformation on social media, this dataset was acquired via the Twitter API. It features a streaming dataset centered on keywords with over 1.8 million tweets, complemented by a historical account-level dataset containing more than 135 million tweets.
\item \textit{Climate Change Tweets Ids} \cite{DVN/5QCCUU_2019}: This dataset houses the tweet IDs for 39.6 million tweets related to climate change. These tweets were sourced from the Twitter API between 21 September 2017 and 17 May 2019 using the Social Feed Manager, based on specific search keywords.
\end{itemize}

\paragraph{Comprehension and Reasoning} Datasets geared towards comprehension and generation typically provide consistent contextual materials, guiding LLMs in the regeneration or continuation of text.

\begin{itemize}
\item \textit{Stanford Question and Answer Dataset (SQuAD)} \cite{rajpurkar2016squad}: This reading comprehension dataset features 100k Q\&A pairs, encompassing subjects from musical celebrities to abstract notions. It draws samples from the top 10k English Wikipedia articles sourced via PageRank. From this collection, 536 articles were randomly selected, excluding passages shorter than 500 words. Crowdsource contributors frame the questions based on these experts, while additional personnel provide the answers.
\item \textit{SciGen} \cite{moosavi2021scigen}: This task centers on reasoning from perceptual data to generate text. It comprises tables from scientific articles alongside their descriptions. The entire dataset is sourced from the ``Computer Science'' section in the arXiv website, holding up to 17.5k samples from ``Computation and Language'' and another 35.5k from domains like ``Machine Learning''. Additionally, the dataset facilitates the evaluation of generative models' arithmetic reasoning capabilities using intricate input formats, such as scientific tables.
\item \textit{ROCStories Corpora (ROC)} \cite{mostafazadeh2016corpus}: Aimed at natural language understanding, this dataset tasks systems with determining the apt conclusion to a four-sentence narrative. It is a curated collection of 50k five-sentence stories reflecting everyday experiences. Beyond its primary purpose, it can also support tasks like story generation.
\item \textit{HellaSwag} \cite{zellers2019hellaswag}: Focused on common-sense reasoning, this dataset encompasses approximately 70k questions. Utilizing Adversarial Filtering (AF), the dataset crafts misleading and intricate false answers for multiple-choice settings, where the objective is to pinpoint the correct answer in context.
\end{itemize}

\subsection{Evaluation Benchmarks}
\label{evaluation_benchmarks}

\begin{table}[!ht]
\caption{Summary of  benchmarks for LLM-generated text detection.}
\renewcommand\arraystretch{1.2}
\fontsize{8}{11}\selectfont
\centering
\resizebox{\textwidth}{!}{
\begin{tabular}{ p{3.7cm} p{0.7cm} p{0.6cm} p{0.6cm} p{3.0cm} p{1.4cm} p{1.3cm} p{2.3cm} }
\toprule
Corpus & Use & Human & LLMs & LLMs Type & Language & Attack & Domain \\
\midrule
TuringBench \cite{uchendu2021turingbench} & train & \textasciitilde8k & \textasciitilde159k & GPT-1, GPT-2, GPT-3, GROVER, CTRL, XLM, XLNET, FAIR, TRANSFORMER\_XL, PPLM & English & - & News Writing \\
MGTBench \cite{he2023mgtbench} & train test & \textasciitilde2.4k \textasciitilde0.6k & \textasciitilde14.4k \textasciitilde3.6k & ChatGPT, ChatGPT-turbo, ChatGLM,
Dolly, GPT4All, StableLM & English & Adversarial & Scientific Writing, Story Generation, News Writing \\
GPABenchmark \cite{liu2023check} & test & \textasciitilde150k & \textasciitilde450k & GPT-3.5 & English & Paraphrase & Scientific Writing \\
Scientific-articles Benchmark \cite{mosca2023distinguishing} & test & \textasciitilde16k & \textasciitilde13k & SCIgen, GPT-2, GPT-3, ChatGPT, Galactica & English & - & Scientific Writing \\
MULTITuDE~\cite{DBLP:conf/emnlp/MackoMULYPSL0SB23} & train test & \textasciitilde4k \textasciitilde3k & \textasciitilde40k \textasciitilde26k & Alpaca-lora, GPT-3.5-Turbo, GPT-4, LLaMA, OPT, OPT-IML-Max, Text-Davinci-003, Vicuna & Arabic, Catalan, Chinese, Czech, Dutch, English, German, Portuguese, Russian, Spanish, Ukrainian & - & Scientific Writing, News Writing, Social Media  \\
HANSEN~\cite{DBLP:journals/corr/abs-2310-16746} & test & - & \textasciitilde21k & ChatGPT, PaLM2, Vicuna13B & English & - & Spoken Text \\
M4 \cite{wang2023m4} & train valid test & \textasciitilde35k \textasciitilde3.5k \textasciitilde3.5k & \textasciitilde112k \textasciitilde3.5k \textasciitilde3.5k & GPT-4, ChatGPT, GPT-3.5, Cohere, Dolly-v2, BLOOMz 176B & English, Chinese, Russian, Urdu, Indonesian, Bulgarian, Arabic & - & Web Text, Scientific Writing, News Writing, Social Media, QA\\
\bottomrule
\end{tabular}
}
\label{tab:benchmark}
\end{table}
\textcolor{black}{Benchmarks with high quality can help researchers verify whether their detectors are feasible and effective rapidly. We sort out and compare the benchmarks that are currently popular or with potential, as shown in \autoref{tab:benchmark}. On the one hand, we hope to help researchers better understand their differences to choose suitable benchmarks for their experiments. On the other hand, we hope to draw researchers' attention to the latest benchmarks, which have been fully designed to verify the latest issues for the task, with great potential.}

\paragraph{TuringBench}
The TuringBench dataset \cite{uchendu2021turingbench} is an initiative designed to explore the challenges of the ``Turing test'' in the context of neural text generation techniques. It comprises human-written content derived from 10k news articles, predominantly from reputable sources such as CNN. For the purpose of this dataset, only articles ranging between 200 to 400 words were selected. LLM-generated text within this dataset is produced by 19 distinct text generation models, including GPT-1, GPT-2 variants (small, medium, large, xl, and pytorch), GPT-3, different versions of GROVER (base, large, and mega), CTRL, XLM, XLNET variants (base and large), FAIR for both WMT19 and WMT20, Transformer-XL, and both PLM variants (distil and GPT-2). Each model contributed ~8k samples, categorized by label type. Notably, TuringBench emerged as one of the pioneering benchmark environments for the detection of LLM-generated text. However, given the rapid advancements in LLM technologies, the samples within TuringBench are now less suited for training and validating contemporary detector performances. As such, timely updates incorporating the latest generation models and their resultant texts are imperative.

\paragraph{MGTBench}
Introduced by \citet{he2023mgtbench}, MGTBench stands as the inaugural benchmark framework for MGT detection. It boasts a modular architecture, encompassing an input module, a detection module, and an evaluation module. The dataset draws upon several of the foremost LLMs, including ChatGPT, ChatGLM, Dolly, ChatGPT-turbo, GPT4All, and StableLM, for text generation. Furthermore, it incorporates over ten widely-recognized detection algorithms, demonstrating significant potential.

\paragraph{GPABenchmark}
The GPABenchmark \cite{liu2023check} is a comprehensive dataset encompassing 600k samples. These samples span human-written, GPT-written, GPT-completed, and GPT-polished abstracts from a broad spectrum of academic disciplines, such as computer science, physics, and the humanities and social sciences. This dataset meticulously captures the quintessential scenarios reflecting both the utilization and potential misapplication of LLMs in academic composition. Consequently, it delineates three specific tasks: generation of text based on a provided title, completion of a partial draft, and refinement of an existing draft. Within the domain of academic writing detection, GPABenchmark stands as a robust benchmark, attributed to its voluminous data and its holistic approach to scenario representation.

\paragraph{Scientific-articles Benchmark}
The Scientific-articles Benchmark \cite{mosca2023distinguishing} comprises 16k human-written articles alongside 13k LLM-generated samples. The human-written articles are sourced from the arXiv dataset available on Kaggle. In contrast, the machine-generated samples, which include abstracts, introductions, and conclusions, are produced by SCIgen, GPT-2, GPT-3, ChatGPT, and Galactica using the titles of the respective scientific articles as prompts. A notable limitation of this dataset is its omission of various adversarial attack types.

\textcolor{black}{
\paragraph{MULTITuDE}
It is a benchmark for detecting machine-generated text in multiple languages. This dataset consists of 74k machine-generated texts and 7k human-written texts across 11 languages \cite{DBLP:conf/emnlp/MackoMULYPSL0SB23}, including Arabic, Catalan, Chinese, Czech, Dutch, English, German, Portuguese, Russian, Spanish, and Ukrainian. The machine-generated texts are produced by eight generative models, including Alpaca-Lora, GPT-3.5-turbo, GPT-4, LLaMA, OPT, OPT-IML-Max, Text-Davinci-003, and Vicuna. In an era of rapidly increasing numbers of multilingual LLMs, MULTITuDE serves as an effective benchmark for assessing the detection capabilities of LLM-generated text detectors in various languages.
}

\textcolor{black}{
\paragraph{HANSEN}
The Human and AI Spoken Text Benchmark (HANSEN) \cite{DBLP:journals/corr/abs-2310-16746} is the largest benchmark for spoken text, encompassing the organization of 17 speech datasets and records, as well as 23k novel AI-generated spoken texts. The AI-generated spoken texts in HANSEN were created by ChatGPT, PaLM2, and Vicuna-13B. Due to the stylistic differences between spoken and written language, detectors may require a more nuanced understanding of spoken text. HANSEN can effectively assess the progress in research aimed at developing such nuanced detectors.
}

\paragraph{M4}
M4 \cite{wang2023m4} serves as a comprehensive benchmark corpus for the detection of text generated by LLMs. It spans a variety of generators, domains, and languages. Compiled from diverse sources, including wiki pages from various regions, news outlets, and academic portals, the dataset reflects common scenarios where LLMs are utilized in daily applications. The LLM-generated texts in M4 are created using cutting-edge generative models such as ChatGPT, LLaMa, BLOOMz, FlanT5, and Dolly. Notably, the dataset captures cross-lingual subtleties, featuring content in more than ten languages. In summary, while the M4 dataset proficiently tackles complexities across domains, models, and languages, it could be further enriched by incorporating a broader range of adversarial scenarios.

\subsection{Data Challenges}

In light of our extensive experience in the area, a notable deficiency persists in the realm of robust datasets and benchmarks tailored for LLMs. Despite commendable advancements, current efforts remain insufficient. A noticeable trend among researchers is the tendency to utilize datasets originally designed for other tasks as human-written texts, and produce LLM-generated texts base on them for training detectors. This approach arises from the limitations of existing datasets or benchmarks in comprehensively addressing diverse research perspectives. As a result, we aim to outline the prominent limitations and challenges associated with the current datasets and benchmarks in this article.

\subsubsection{Comprehensiveness of Evaluation Frameworks}
Before gaining trust, a reliable detector demands multifaceted assessment. The current benchmarks are somewhat limited, providing only superficial challenges and thereby not facilitating a holistic evaluation of detectors. We highlight five crucial dimensions that are essential for the development of more robust benchmarks for LLM-generated text detection task. These dimensions include the incorporation of multiple types of attacks, diverse domains, varied tasks, a spectrum of models, and the inclusion of multiple languages.

\paragraph{Multiple Types of Attack} are instrumental in ascertaining the efficacy of detection methodologies. In practical environments, LLM-generated text detectors often encounter texts that are generated using a wide range of attack mechanisms, which differ from texts generated through simple prompts.
For instance, the \textit{prompt attack} elucidated in \autoref{potential_attacks} impels the generative model to yield superior-quality text, leveraging intricate and sophisticated prompts. Integrating such texts into prevailing datasets is imperative. This concern is also echoed in the limitations outlined by \citet{guo2023close}.

\paragraph{Multi-domains and multi-tasks} configurations are pivotal in assessing a detector's performance across diverse real-world domains and LLM applications. These dimensions bear significant implications for a detector's robustness, usability, and credibility. For instance, in scholarly contexts, a proficient detector should consistently excel across all fields. In everyday scenarios, it should adeptly identify LLM-generated text spanning academic compositions, news articles, arithmetic reasoning, and Q\&A sessions. While numerous existing studies prudently incorporate these considerations, we advocate for the proliferation of superior-quality datasets.

\paragraph{Multiple LLMs} The ongoing research momentum in LLMs has ushered in formidable counterparts like LLaMa \cite{llama}, PaLM \cite{palm}, and Claude-2,\footnote{\url{https://www.anthropic.com/index/claude-2}} rivaling ChatGPT's prowess. As the spotlight remains on ChatGPT, it is essential to concurrently address potential risks emanating from other emerging LLMs.

\paragraph{Multilingual} considerations demand increased attention. We strongly encourage researchers to spearhead the creation of multilingual datasets to facilitate the evaluation of text detectors generated by LLMs across different languages. The utilization of pre-trained models may uncover instances where certain detectors struggle with underrepresented languages, while LLMs could exhibit more noticeable inconsistencies. This dimension presents a rich avenue for exploration and discourse.

\subsubsection{Temporal}
It is discernible that certain contemporary studies persistently employed seminal but somewhat antiquated benchmark datasets, which had significantly shaped prior GPT-generated text and fake news detection endeavors. However, these datasets predominantly originate from backward LLMs, implying that validated methodologies might not invariably align with current real-world dynamics. We emphasize the significance of utilizing datasets formulated with advanced and powerful LLMs, while also urging benchmark dataset developers to regularly update their contributions to reflect the rapid evolution of the field.
\begin{figure}
\centering
\begin{forest}
  for tree={
  grow=east,
  reversed=true,
  anchor=base west,
  parent anchor=east,
  child anchor=west,
  base=left,
  font=\scriptsize,
  rectangle,
  draw,
  rounded corners,align=left,
  minimum width=2.12em,
  inner xsep=4pt,
  inner ysep=1pt,
  },
  where level=1{font=\scriptsize,fill=pink!5}{},
  where level=2{font=\tiny,yshift=0.26pt,fill=yellow!20}{},
    [Advanced \\ Detector \\ Research\\(Sec. 5), text width=3.2em, fill=blue!10,
        [{\includegraphics[width=4.3cm]{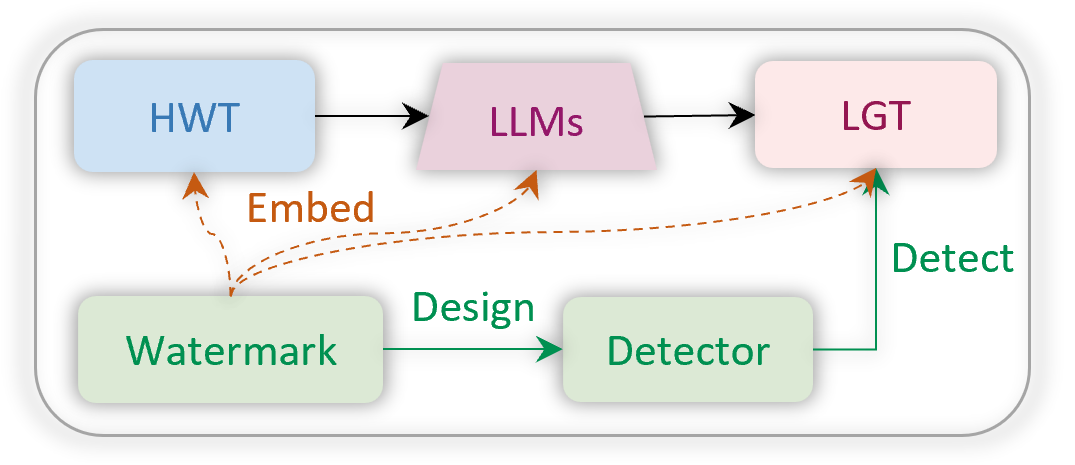}} \\ Watermarking Technology, text width=12em
            [\emph{Data-Driven Watermarking:} \cite{gu2022watermarking} / \cite{lucas2023gpts} / \\ \cite{DBLP:journals/corr/abs-2303-11470}, text width=19em]
            [\emph{Model-Driven Watermarking:} \cite{kirchenbauer2023watermark} / \cite{DBLP:journals/corr/abs-2305-15060} / \\ \cite{kirchenbauer2023reliability} / \cite{DBLP:journals/corr/abs-2310-06356} / \cite{liu2023private} / \\ \cite{DBLP:journals/corr/abs-2307-15593} / \cite{DBLP:journals/corr/abs-2310-03991}, text width=19em]
            [\emph{Post-Processing Watermarking:} \cite{DBLP:journals/jss/PorWC12} / \\ \cite{DBLP:conf/ideas/RizzoBM16} / \cite{DBLP:conf/mmsec/TopkaraTA06} / \\ \cite{DBLP:conf/aaai/YangZCZMWY22} / \cite{DBLP:journals/corr/abs-2305-05773} / \cite{DBLP:conf/acl/YooAJK23} / \\ \cite{DBLP:journals/corr/abs-2305-08883} / \cite{DBLP:conf/sp/AbdelnabiF21} / \cite{DBLP:journals/corr/abs-2310-12362}, text width=19em]
        ]
        [{\includegraphics[width=4.3cm]{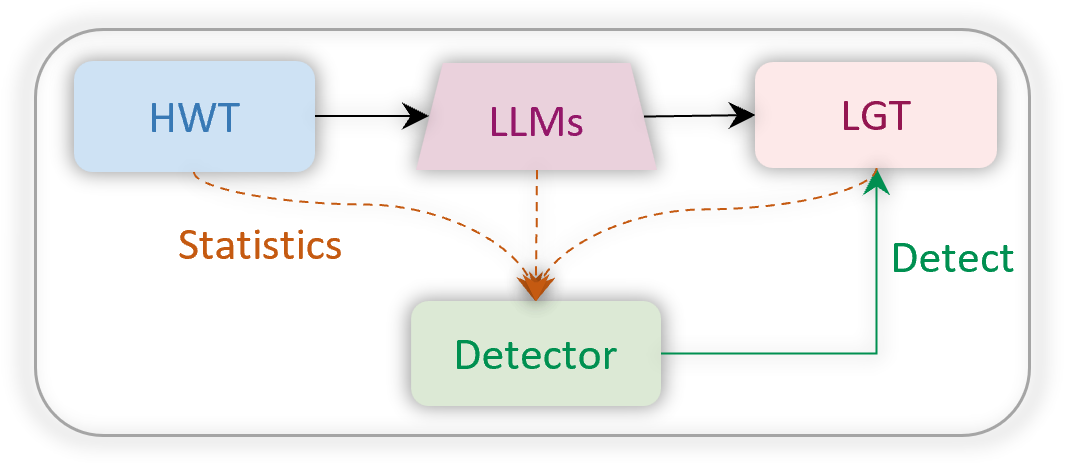}} \\ Statistics-Based Detectors, text width=12em
            [\emph{Linguistics Features Statistics:} \cite{corston2001machine} / \\ \cite{kalinichenko2003digital} / \cite{baayen2001word} / \cite{arase2013machine} / \\ \cite{DBLP:journals/corr/abs-2111-02878} / \cite{DBLP:journals/corr/abs-2308-11767}, text width=19em]
            [\emph{White-Box Statistics:} \cite{solaiman2019release} / \cite{gehrmann2019gltr} \\ \cite{su2023detectllm} / \cite{lavergne2008detecting} / \cite{beresneva2016computer} / \\ \cite{vasilatos2023howkgpt} / \cite{DBLP:conf/emnlp/WuPSCC23} / \cite{mitchell2023detectgpt} / \cite{deng2023efficient} / \\ \cite{bao2023fast} / \cite{DBLP:journals/corr/abs-2305-17359} / \cite{tulchinskii2023intrinsic}, text width=19em]
            [\emph{Black-Box Statistics:} \cite{DBLP:journals/corr/abs-2305-17359} / \cite{DBLP:journals/corr/abs-2401-12970} / \cite{DBLP:conf/emnlp/ZhuYCCFHDL0023} / \\ \cite{yu2023gpt} / \cite{DBLP:conf/bea/QuidwaiLD23} / \cite{DBLP:journals/corr/abs-2311-07700}, text width=19em]
        ]
        [{\includegraphics[width=4.3cm]{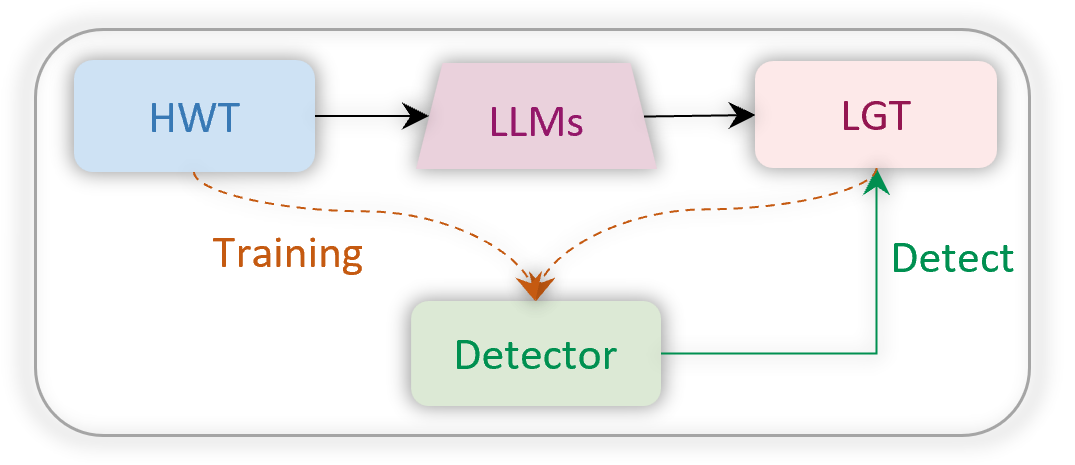}} \\ Neural-Based Detectors, text width=12em 
            [\emph{Feature-Based Classifiers:} \cite{DBLP:conf/coling/AichBP22} / \cite{shah2023detecting} / \\ \cite{DBLP:conf/ijcnn/CorizzoL23} / \cite{DBLP:journals/corr/abs-2308-05341} / \\ \cite{DBLP:conf/icnlsp/SchaaffSM23} / \cite{DBLP:journals/coling/SchusterSSB20} / \\ \cite{DBLP:conf/ijcnn/CrothersJVB22} / \cite{DBLP:journals/corr/abs-2304-14072} / \cite{DBLP:journals/corr/abs-2310-08903} / \cite{wu2023mfd}, text width=19em 
            ]
            [\emph{Pre-Training Classifiers:} \cite{bakhtin2019real} /  \cite{uchendu2020authorship} / \\ \cite{antoun2023robust} / \cite{li2023deepfake} / \cite{Fagni_2021} / \cite{gambini2022pushing} / \\ \cite{guo2023close} / \cite{liu2023argugpt} / \cite{liu2023check} / 
            \cite{DBLP:journals/corr/abs-2306-07401} / \\
            \cite{DBLP:journals/corr/abs-2306-07401} / \cite{bakhtin2019real} / \cite{uchendu2020authorship} / \\ \cite{antoun2023robust} / \cite{li2023deepfake} / \cite{sarvazyan2023supervised} / \\ \cite{DBLP:conf/naacl/RodriguezHGSS22} / \cite{liu2022coco} / \cite{zhong2020neural} / \\\cite{conda} / \cite{yang2023chatgpt} /  \cite{shi2023red} / \\ \cite{koike2023outfox} / \cite{he2023mgtbench} / \cite{hu2023radar} / \\\cite{koike2023outfox} / \cite{DBLP:journals/corr/abs-2304-14106} /  \cite{DBLP:journals/corr/abs-2309-03164} / \\ \cite{DBLP:conf/emnlp/CowapGF23} / \cite{DBLP:journals/corr/abs-2309-12934} 
            , text width=19em 
            ]
            [\emph{LLMs as Detectors:} \cite{zellers2019defending} / \cite{liu2023argugpt} / \\ \cite{bhattacharjee2023fighting} / \cite{koike2023outfox}, text width=19em 
            ]
        ]
        [{\includegraphics[width=4.3cm]{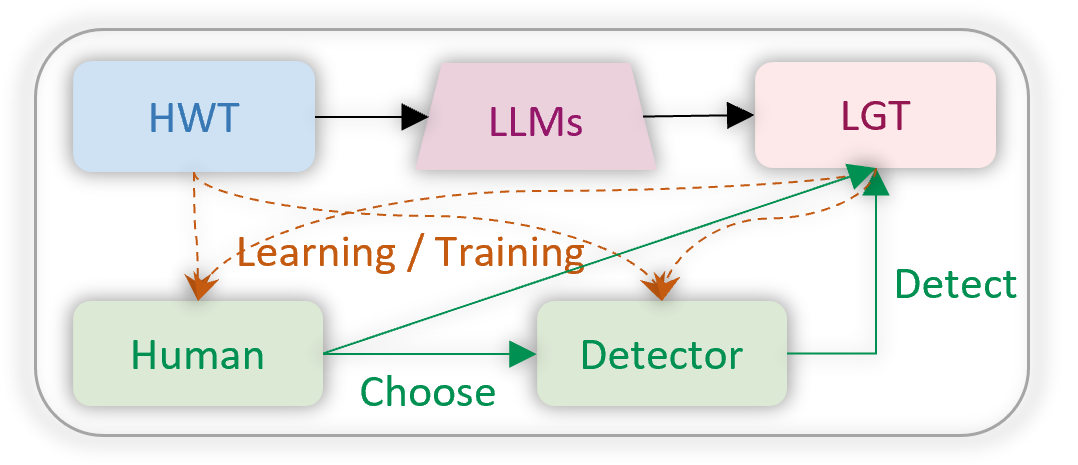}} \\ Human-Assisted Methods, text width=12em 
            [\emph{Intuitive Indicators:} \cite{uchendu2022does} / \cite{dugan2023real} / \\ / \cite{jawahar2020automatic}, text width=19em]
            [\emph{Imperceptible Features:} \cite{ippolito2019automatic} / \cite{DBLP:conf/acl/ClarkASHGS20} / \\ \cite{gehrmann2019gltr}, text width=19em]
            [\emph{Enhancing Human Detection Capabilities:} \cite{ippolito2019automatic} / \cite{dugan2020roft} / \\ \cite{dugan2023real} / \cite{dou2021gpt}, text width=19em]
            [\emph{Mixed Detection: Understanding and Explanation:} \cite{weng2023towards}, text width=19em]
        ]
    ]
\end{forest}
\caption{
\textcolor{black}{Classification of LLM-generated text detectors with corresponding diagrams and paper lists. We categorize the detectors into watermarking technology, statistics-based detectors, neural-based detectors, and human-assisted methods. In the diagrams, HWT represents Human-Written Text and LGT represents LLM-Generated Text. We use the \textcolor{orange}{orange} lines to highlight the source of the detector's detection capability, and the \textcolor{green}{green} lines to describe the detection process.}
}
\label{fig:detectors}
\end{figure}

\section{Advances in Detector Research}
\label{detectors}

In this section, we present different detector designs and detection algorithms, including watermarking technology, \textcolor{black}{statistics-based detectors}, \textcolor{black}{neural-based detectors}, and human-assisted methods. We focus on the most recently proposed methods and divide our discussion according to their underlying principles (see \autoref{fig:detectors}).

\subsection{Watermarking Technology}

Originally deployed within the realm of computer vision for the development of generative models, watermarking techniques have been integral to the detection of AI-generated images, serving as protective measures for intellectual and property rights in the visual arts. With the advent and subsequent proliferation of LLMs, the application of watermarking technology has expanded to encompass the identification of text generated by these models. Watermarking techniques not only protect substantial models from unauthorized acquisition, such as through sequence distillation but also mitigate the risks associated with replication and misuse of LLM-generated text.

\textcolor{black}{\subsubsection{Data-Driven Watermarking} Data-driven methods enable the verification of data ownership or the tracking of illegal copying or misuse by embedding specific patterns or tags within the training datasets of LLMs. These methods typically rely on backdoor insertion, where a small number of watermarked samples are added to the dataset, allowing the model to implicitly learn a secret function set by the defender. When a specific trigger is activated, the backdoor watermark is triggered, which is usually implemented in a black-box setting \cite{gu2022watermarking}. This mechanism protects the model from unauthorized fine-tuning or use beyond the terms of the license by embedding a backdoor during the foundational and multi-task learning framework phases of model training, specified by the owner's input. Even if the model is fine-tuned for several downstream tasks, the watermark remains difficult to eradicate.
}

\textcolor{black}{
However, subsequent studies identified vulnerabilities in this technology, showing that it can be relatively easily compromised. \citet{lucas2023gpts} detailed an attack method on this watermarking strategy by analyzing the content generated by autoregressive models to precisely locate the trigger words or phrases of the backdoor watermark. The study points out that triggers composed of randomly combined common words are easier to detect than those composed of unique and rare markers. Additionally, the research mentions that access to the model's weights is the only prerequisite for detecting the backdoor watermark. Recently, \citet{DBLP:journals/corr/abs-2303-11470} introduced a clean-label backdoor watermarking framework that uses subtle adversarial perturbations to mark and trigger samples. This method effectively protects the dataset while minimizing the impact on the performance of the original task. The results show that adding just 1\% of watermarked samples can inject a traceable watermark feature.
}

\textcolor{black}{
It is important to note that data-driven methods were initially designed to protect copyright of datasets and therefore generally lack substantial payload capacity and generalizability. Moreover, applying such techniques in the field of LLM-generated text detection requires significant resource investment, including the embedding of watermarks in a vast amount of data and the retraining of LLMs.}
|

\textcolor{black}{\subsubsection{Model-Driven Watermarking} Model-Driven methods embed watermarks directly into the LLMs by manipulating the logits output distribution or token sampling during the inference process. As a result, the LLMs generate responses that carry the embedded watermark.}

\textcolor{black}{\paragraph{Logits-Based Methods} \citet{kirchenbauer2023watermark} were the first to design a logits-based watermarking framework for LLMs, characterized by minimal impact on text quality. This framework facilitates the detection process through efficient open-source algorithms, eliminating the need to access the LLM's API or parameters. Before text generation, the method randomly selects a set of ``green'' tokens, defines the rest as ``red,'' and then gently guides the model to choose tokens from the ``green'' set during sampling. Additionally, \citet{kirchenbauer2023watermark} developed a watermark detection method based on interpretable $p$-values, which identifies watermarks by performing statistical analysis on the red and green tokens within the text to calculate the significance of the $p$-values. Following \cite{kirchenbauer2023watermark}, \citet{DBLP:journals/corr/abs-2305-15060} introduced a new watermarking method called SWEET, which elevates ``green'' tokens only at positions with high token distribution entropy during the generation process, thus maintaining the watermark's stealth and integrity. It uses entropy-based statistical tests and Z-scores for detecting watermarked code.}

\textcolor{black}{Despite the excellent performance of \citet{kirchenbauer2023watermark}, its robustness is still debated. Recent work from \citet{kirchenbauer2023reliability} studies the resistance of watermarked texts to attacks via manual rewriting, rewriting using an unwatermarked LLM, or integration into a large corpus of handwritten documents. This study introduces a window testing method called ``WinMax'' to evaluate the effectiveness of accurately detecting watermarked areas within a large number of documents. To address the challenges of synonym substitution and text paraphrasing, \citet{DBLP:journals/corr/abs-2310-06356} proposed a semantic invariant robust watermarking method for LLMs. This method generates semantic embeddings for all preceding tokens and uses them to determine the watermark logic, demonstrating robustness to synonym substitution and text paraphrasing. Moreover, current watermark detection algorithms require a secret key during generation, which can lead to security vulnerabilities and forgery in public detection processes. To address this issue, \citet{liu2023private} introduced the first dedicated private watermarking algorithm for watermark generation and detection, deploying two different neural networks for each stage. By avoiding the use of the secret key in both stages, this method innovatively extends existing text watermark algorithms. Furthermore, it shares certain parameters between the watermark generation and detection networks, thus improving the efficiency and accuracy of the detection network while minimizing the impact on the speed of both generation and detection processes.}

\textcolor{black}{\paragraph{Token Sampling-Based Methods} During the normal model inference process, token sampling is determined by the sampling strategy and is often random, which helps guide the LLMs to produce more unpredictable text. Token sampling-based methods achieve watermarking by influencing the token sampling process, either by setting random seeds or specific patterns for token sampling. \citet{DBLP:journals/corr/abs-2307-15593} employed a sequence of random numbers as a secret watermark key to intervene in and determine the token sampling, which is then mapped into the LLMs to generate watermarked text. In the detection phase, the secret key is utilized to align the text with the random number sequence for the detection task. The method demonstrates strong robustness against paraphrasing attacks, even when approximately 40-50\% of the tokens have been modified.}

\textcolor{black}{Another recent work is SemStamp \cite{DBLP:journals/corr/abs-2310-03991}, a robust sentence-level semantic watermarking algorithm based on Locality-Sensitive Hashing (LSH). This algorithm starts by encoding and LSH hashing the candidate sentences generated by the LLM, dividing the semantic embedding space into watermarked and non-watermarked regions. It then continuously performs sentence-level rejection sampling until a sampled sentence falls into the watermarked partition of the semantic embedding space. Experimental results indicate that this method is not only more robust than previous SOTA methods in defending against common and more effective bigram paraphrase attacks but also superior in maintaining the quality of text generation.}

\textcolor{black}{In general, model-driven watermarking is a plug-and-play method that does not require any changes to the model's parameters and has minimal impact on text quality, making it a reliable and practical watermarking approach. However, there is still significant opportunity for improvement in its robustness, and its specific usability needs to be further explored through additional experiments and practical applications.}

\textcolor{black}{
\subsubsection{Post-Processing Watermarking} 
Post-processing watermarking refers to a technique that involves embedding a watermark by processing the text after it has been output by a LLM. This method typically functions as a separate module that works in a pipeline with the output of the generative model.
}

\textcolor{black}{
\paragraph{Character-Embedded Methods}
Early post-processing watermarking techniques relied on the insertion or substitution of special Unicode characters into text. These characters are difficult for the naked eye to recognize but carry distinct encoding information \cite{DBLP:journals/jss/PorWC12, DBLP:conf/ideas/RizzoBM16}. More recently, \citet{DBLP:conf/ideas/RizzoBM16} introduced Easymark, a method which ingeniously utilizes the fact that Unicode has many code points with identical or similar appearances. Specifically, Easymark embeds watermarks by replacing the regular space character (U+0020) with another whitespace code point (e.g., U+2004), using Unicode's variation selectors, substituting substrings, or using spaces and homoglyphs of slightly different lengths, all while ensuring that the appearance of the text remains virtually unchanged. The results indicate that watermarks embedded by Easymark can be reliably detected without reducing the BLEU score or increasing perplexity of the text, surpassing existing advanced techniques in terms of both quality and watermark reliability.
}

\textcolor{black}{
\paragraph{Synonym Substitution-Based Methods}
In light of the vulnerability of character-level methods to targeted attacks, some research has shifted towards embedding watermarks at the word level, mainly through synonym substitution. Early watermark embedding schemes involve the continuous replacement of words with synonyms until the text carries the intended watermark content. To address this, \citet{DBLP:conf/mmsec/TopkaraTA06} introduced a quantifiable and resilient watermarking technique using Wordnet~\cite{fellbaum1998wordnet}. Building upon this, \citet{DBLP:conf/aaai/YangZCZMWY22, DBLP:journals/corr/abs-2305-05773, DBLP:conf/acl/YooAJK23} employed pre-trained or further fine-tuned neural models to perform word replacement and detection tasks, thereby better preserving the semantic integrity of the original sentences. Additionally, \citet{DBLP:journals/corr/abs-2305-08883} defined a binary encoding function to calculate random binary codes corresponding to words, and selectively replaced words representing a binary ''0`` with contextually relevant synonyms representing a binary ''1``, effectively embedding the watermark. Experiments have demonstrated that this method ensures the watermark's robustness against attacks such as retranslation, text polishing, word deletion, and synonym substitution without compromising the original text's semantics.
}

\textcolor{black}{
\paragraph{Sequence-to-Sequence Methods} Recent research has explored end-to-end watermark encryption techniques with the goal of enhancing flexibility and reducing the presence of artifacts introduced by watermarks. For instance, \citet{DBLP:conf/sp/AbdelnabiF21} proposed Adversarial Watermark Transformer (AWT), the first framework to automate the learning of word replacements and their contents for watermark embedding. This method combines end-to-end and adversarial training, capable of injecting binary messages into designated input text at the encoding layer, producing an output text that is unnoticeable and minimally alters the semantics and correctness of the input. The method employs a transformer encoder layer to extract secret messages embedded within the text. Similarly, \citet{DBLP:journals/corr/abs-2310-12362} introduced the REMARK-LLM framework, which includes three components: (i) a message encoding module that injects binary signatures into texts generated by LLMs; (ii) a reparametrization module that converts the dense distribution of message encoding into a sparse distribution for generating watermarked text tokens; (iii) a decoding module dedicated to extracting signatures. Experiments suggest that REMARK-LLM embeds more signature bits into the same text while maintaining semantic integrity and showing enhanced resistance to various watermark removal and detection attacks compared to AWT.
}

\textcolor{black}{
Compared to model-driven watermarking, post-processing watermarking may depend more heavily on specific rules, making it more vulnerable to sophisticated attacks that exploit visible clues. Despite this risk, post-processing watermarking has significant potential for various applications. Many existing watermarking techniques typically necessitate training within white-box models, making them unsuitable for use with black-box LLMs setting. For instance, embedding watermarks in GPT-4 is nearly impossible given its proprietary and closed-source nature. Nevertheless, post-processing watermarking provides a solution for adding watermarks to text generated by black-box LLMs, enabling third parties to embed watermarks independently.
}

\subsection{Statistics-Based Methods}

Within the statistics-based setup, this subsection presents methods for proficiently identifying text generated by LLMs using detectors, without the need for additional training through supervised signals. This approach assumes access to LLMs or extract features from text, \textcolor{black}{and is evaluated based on unique features and statistical data to derive statistical regularities (e.g., compute thresholds).}

\subsubsection{Linguistics Features Statistics}
The inception of statistics-based detection research can be traced back to the pioneering work of \citet{corston2001machine}. In this foundational study, the authors utilized linguistic features, such as the branching properties observed in grammatical analyses of text, function word density, and constituent length, to determine whether a given text was generated by a machine translation model. These features served as key indicators in distinguishing machine-generated text from human-generated text.

Another notable method, dedicated to achieving similar detection objectives, employs frequency statistics. For instance, \citet{kalinichenko2003digital} utilized the frequency statistics associated with word pairs present in the text as a mechanism to ascertain whether the text had been autonomously generated by a generative system. Furthermore, the approach adopted by \citet{baayen2001word} is grounded in the distributional features characteristic of words. Progressing in this line of inquiry, \citet{arase2013machine} later contributed by developing a detection technique that captures the ``phrase salad'' phenomenon within sentences.

\textcolor{black}{Recent studies on LLM generated text detection have proposed methodologies based on linguistics features statistics. \citet{DBLP:journals/corr/abs-2111-02878} proposed a method of using repeated high-order \textit{n}-grams to detect LLM-generated documents. This approach is predicated on the observation that certain \textit{n}-grams recur with unusual frequency within LLM-generated text, a phenomenon that has been documented extensively. Similarly, \citet{DBLP:journals/corr/abs-2308-11767} developed a detection system based on statistical similarities of bigrams. Their findings indicate that only 23\% of bigrams in texts generated by ChatGPT are unique, underscoring significant disparities in terminology usage between human and LLM-generated content. Impressively, their algorithm successfully identified 98 of 100 LLM-written academic papers, thereby demonstrating the efficacy of their feature engineering approach in distinguishing LLM-generated texts.
}

\textcolor{black}{
However, our empirical observations reveal a conspicuous limitation in the application of linguistic feature statistics:  the availability of these methods relies heavily on extensive corpus statistics and various types of LLMs.
}

\subsubsection{White-Box Statistics}

\textcolor{black}{Currently, white-box methods for detecting text generated by LLMs require direct access to the source model for implementation. The existing white-box detection techniques primarily use zero-shot approaches, which involves obtaining the model's logits output and calculating specific metrics. These metrics are then compared against predetermined thresholds obtained through statistical methods to identify LLM-generated text.}

\paragraph{Logits-Based Statistics}

\textcolor{black}{Logits are the raw outputs produced by LLMs during text generation, specifically from the model's final linear layer, which is typically located before the softmax function. These outputs indicate the model's confidence levels associated with generating each potential subsequent word. The Log-Likelihood~\cite{solaiman2019release}, a metric derived directly from the logits, measures the average token-wise log probability for each token within the provided text by consulting the originating LLM. This measurement helps to determine the likelihood of the text being generated by an LLM. At present, the Log-Likelihood is recognized as one of the most popular baseline metrics for LLM-generated text detection task.}

\textcolor{black}{Similarly, Rank~\cite{solaiman2019release} is another normal baseline computed from logits. The Rank metric calculates the ranking of each word in a sample within the model's output probability distribution. This ranking is determined by comparing the logit score of the word against the logit scores of all other possible words. If the average rank of each word in the sample is high, it suggests that the sample is likely generated by LLMs. Log-Rank, on the other hand, further processes each token's rank value by applying a logarithmic function and has garnered increasing attention. One noteworthy method based on this intuitive approach is GLTR \cite{gehrmann2019gltr}, which is designed as a visual forensic tool to facilitate comparative judgment. The tool divides different marking colors according to the sampling frequency level of the token, and highlights the proportion of words that LLMs tend to use in the analyzed text by marking different colors for the tokens that provide the text. The Log-Likelihood Ratio Ranking (LRR) proposed by \citet{su2023detectllm} combines Log-Likelihood and Log-Rank by taking the ratio of the two metrics. This approach enhances performance by effectively complementing Log Likelihood assessments with Log Rank analysis to provide a more comprehensive assessment.}

Entropy represents another early zero-shot method used for evaluating LLM-generated text. It is typically employed to measure the uncertainty or the amount of information in a text or model output, and is also calculated through the probability distribution of words. High entropy indicates that the content of the sample text is unclear or highly diversified, meaning that many words have a similar probability of being chosen. In such cases, the sample is likely to have been generated by an LLM. \citet{lavergne2008detecting} employed the Kullback-Leibler (KL) divergence to assign scores to \textit{n}-grams, taking into account the semantic relationships between their initial and final words. This approach identifies \textit{n}-grams with significant dependencies between the initial and terminal words, thus aiding in the detection of spurious content and enhancing the overall performance of the detection process.

The method employing perplexity, grounded in traditional $n$-gram LMs, evaluates the proficiency of LMs at predicting text \cite{beresneva2016computer}. More recent work, such as HowkGPT \cite{vasilatos2023howkgpt}, discerns LLM-generated text, specifically homework assignments, by calculating and comparing perplexity scores derived from student-written and ChatGPT-generated text. Through this comparison, thresholds are established to identify the origin of submitted assignments accurately. Moreover, the widely recognized GPTZero\footnote{\url{https://gptzero.me/}} estimates the likelihood of a review text being generated by LLMs. This estimation is based on a meticulous examination of the text’s perplexity and burstiness metrics. 
\textcolor{black}{In a recent study, \citet{DBLP:conf/emnlp/WuPSCC23} unveiled LLMDet, a tool designed to quantify and catalogue the perplexity scores attributable to various models for selected \textit{n}-grams by computing their next-token probabilities. LLMDet exploits the intrinsic self-watermarking characteristics of text, as evidenced by proxy perplexity, to trace the source of the text and to detefct it accordingly. The tool demonstrates a high classification accuracy of 98.54\%, while also offering computational efficiency compared to fine-tuned RoBERTa. In addition, \citet{DBLP:journals/corr/abs-2310-06202} extract UID-based features by analyzing the token probabilities of articles and then trains a logistic regression classifier to fit the UID characteristics of texts generated by different LLMs, in order to identify the origins of the texts. GHOSTBUSTER~\cite{DBLP:journals/corr/abs-2305-15047} inputs text generated by LLMs into a series of weaker language models to obtain token probabilities, and then conducts a structured search on the combinations of these model outputs to train a linear classifier for distinguishing LLM-generated texts. This detector achieves an average F1 score of 99.0, which is an increase of 41.6 F1 score over previous methods such as GPTZero and DetectGPT.}

\paragraph{Perturbed-Based Methods}
\textcolor{black}{Some white-box statistical (or zero-shot) approaches detect LLM-generated text by comparing the differences in performance metrics after statistical perturbation. \citet{mitchell2023detectgpt} proposed a method to identify text produced by LLMs by analyzing structural patterns in the LLMs' probability functions, specifically in regions of negative curvature. The premise is that LLM-generated text tends to cluster at local log-probability maxima. Detection involves comparing log-probabilities of text against those from the target LLM, using a pre-trained mask-filling model like T5 to create semantically similar text perturbations.}

While innovative and sometimes more effective than supervised methods, DetectGPT has limitations, including potential performance drops if rewrites don't adequately represent the space of meaningful alternatives, and high computational demands, as it needs to score many text perturbations. In response to this challenge, \citet{deng2023efficient} proposed a method that uses a Bayesian surrogate model to select a small number of typical samples for scoring. By interpolating the scores of typical samples to other samples to improve query efficiency, the overhead is reduced by half while maintaining performance. \citet{bao2023fast} reported a method that replaces the perturbation step of DetectGPT with a more efficient sampling step, significantly improving the detection accuracy by about 75\% and increasing the detection speed by 340 times. Unlike DetectGPT, the white-box configuration in DNA-GPT~\cite{DBLP:journals/corr/abs-2305-17359} utilizes large language models such as ChatGPT to continue writing truncated texts instead of employing perturbation settings. It analyzes the differences between the original text and the continued text by calculating probability divergence, achieving a detection performance close to 100\%. DetectLLM \cite{su2023detectllm}, another recent contribution, parallels the conceptual framework of DetectGPT. It employs normalized perturbed log-rank for text detection generated by LLMs, asserting a lower susceptibility to the perturbation model and the number of perturbations compared to DetectGPT.

\paragraph{Intrinsic Dimension Estimation} The study conducted by \citet{tulchinskii2023intrinsic} posited the invariant nature of the competencies exhibited by both human and LLMs within their respective textual domains. The proposed approach involves the construction of detectors utilizing the intrinsic dimensions of manifolds underpinning the embedding set of given text samples. More specifically, the methodology entails computing the average intrinsic dimensionality values for both sets of fluent human-written texts and LLM-generated texts in the target natural language. The ensuing statistical separation between these two sets facilitates the establishment of a separation threshold for the target language, thereby enabling the detection of text generated by LLMs. It is imperative to acknowledge the robustness of this approach across various scenarios, including cross-domain challenges, model shifts, and adversarial attacks. However, its reliability falters when confronted with suboptimal or high-temperature generators.

\textcolor{black}{
\subsubsection{Black-Box Statistics}
Unlike white-box statistical methods, black-box statistical methods utilize a black-box model to calculate specific feature scores of a text without needing access to the logits of the source or surrogate model. \citet{DBLP:journals/corr/abs-2305-17359} employed LLMs to continue writing truncated texts under review and defined human-written versus LLM-generated texts by calculating the $n$-gram similarity between the continuation and the original text. Similarly, \citet{DBLP:journals/corr/abs-2401-12970} and \citet{DBLP:conf/emnlp/ZhuYCCFHDL0023} identified LLM-generated texts by computing the similarity scores between the original texts and their rewritten and revised versions. These approaches are based on the observation that human-written texts tend to trigger more revisions when LLMs are tasked with rewriting and editing than LLM-generated texts. \citet{yu2023gpt} introduced a detection mechanism that also capitalizes on the similarity between the original text and the regenerated text. Differing from other methods, this approach initially identifies the original question that prompted the generation of the text and regenerates the text based on this inferred question. Additionally, \cite{DBLP:conf/bea/QuidwaiLD23} analyzes sentences from LLM-generated texts and their paraphrases, distinguishing them from human-written texts by calculating cosine similarity, achieving an accuracy of up to 94\%. \citet{DBLP:journals/corr/abs-2311-07700} introduced a denoising-based black-box zero-shot statistics method that employs a black-box LLM to denoise artificially added noise to input texts. The denoised texts are then semantically compared to the original texts, resulting in an AUROC score of 91.8\%. 
}

\textcolor{black}{However, the approaches of black-box statistics are not without challenges, including the substantial overhead of accessing the LLM and long response times.}

\subsection{Neural-Based Methods}
\subsubsection{Features-Based Classifiers}
\paragraph{Linguistic Features-Based classifiers}
When comparing texts generated by LLMs with those written by humans, the differences in numerous linguistic features provide a solid basis for feature-based classifiers to effectively distinguish between them. The workflow of such classifiers typically starts with the extraction of key statistical language features, followed by the application of machine learning techniques to train a classification model. This approach has been widely used in the identification of fake news. For instance, in the recent study, \citet{DBLP:conf/coling/AichBP22} achieved an impressive accuracy of 97\%by extracting 21 textual features and employing a KNN classifier. Drawing inspiration from the tasks of detecting fake news and LLM-generated texts, the linguistic features of texts can be extensively categorized into stylistic features, complexity features, semantic features, psychological features, and knowledge-based features. These features are primarily obtained through statistical methods.

\textcolor{black}{Stylistic Features primarily focus on the frequency of words that can highlight the stylistic elements of the text, including the frequency of capitalized words, proper nouns, verbs, past tense words, stopwords, technical words, quotes, and punctuation \cite{horne2017just}. Complexity Features are extracted to showcase the complexity of the text, such as the type-token ratio (TTR) and textual lexical diversity (MTLD) \cite{mccarthy2005assessment}. Semantic Features includes Advanced Semantic (AdSem), Lexico Semantic (LxSem), and statistics of semantic dependency tags, among other semantic-level features. These can be extracted using tools like LingFeat \cite{DBLP:conf/emnlp/LeeJL21}. Psychological Features generally related to sentiment analysis, these can be based on tools like SentiWordNet~\cite{DBLP:conf/lrec/BaccianellaES10} to calculate sentiment scores or extracted using sentiment classifiers. Information Features include named entities (NE), opinions (OP), and entity relation extraction (RE), and can be extracted using tools such as UIE~\cite{DBLP:conf/acl/0001LDXLHSW22} and CogIE~\cite{DBLP:conf/acl/JinCSWXZ21}.}

\textcolor{black}{\citet{shah2023detecting} constructed a classifier based on stylistic features such as syllable count, word length, sentence structure, frequency of function word usage, and punctuation ratio. This classifier achieved an accuracy of 93\%, which effectively demonstrates the significance of stylistic features for LLM-generated text detection. Other work integrated text modeling with a variety of linguistic features through data fusion techniques \cite{DBLP:conf/ijcnn/CorizzoL23}, which included different types of punctuation marks, the use of the Oxford comma, paragraph structures, average sentence length, the repetitiveness of high-frequency words, and sentiment scores. On English and Spanish datasets, this approach achieved F1-Scores of 98.36\% and 98.29\% respectively, indicating its exceptional performance. \citet{DBLP:journals/corr/abs-2308-05341} further employed a multidimensional approach to enhance the classifier's discriminative power, which included complexity measures, semantic analysis, list searches, error-based features, readability assessments, artificial intelligence feedback, and text vector features. Ultimately, the optimized detector's performance exceeded that of GPTZero by 183.8\% in F1 Score, showcasing its superior detection capabilities.}

\textcolor{black}{Although classifiers based on linguistic features have their advantages in distinguishing between human and AI-generated texts, their shortcomings cannot be overlooked. The results from \cite{DBLP:conf/icnlsp/SchaaffSM23} indicate that such feature classifiers have poor robustness against ambiguous semantics and often underperform neural network features. Moreover, classifiers based on stylistic features may be capable of differentiating between texts written by humans and those generated by LLMs, but their ability to detect LLM-generated misinformation is limited. This limitation is highlighted in \cite{DBLP:journals/coling/SchusterSSB20}, which shows that language models tend to produce stylistically consistent texts. However, \citet{DBLP:conf/ijcnn/CrothersJVB22} suggests that statistical features can offer additional adversarial robustness and can be utilized in constructing integrated detection models.}

\textcolor{black}{\paragraph{Model Features-Based Classifiers}
In addition to linguistic features, classifiers based on model features have recently garnered considerable attention from researchers. These classifiers are not only capable of detecting texts generated by LLMs but can also be employed for text provenance tracing. Sniffer~\cite{DBLP:journals/corr/abs-2304-14072} involves extracting aligned token-level perplexity and contrastive features, which measure the percentage of words with lower perplexity when comparing one model $\theta_i$ with another model $\theta_j$. By training a linear classifier with these features, an accuracy of 86.0\% was achieved. SeqXGPT~\cite{DBLP:journals/corr/abs-2310-08903} represents further exploration in the field of text provenance tracing, building on the proposed features to design a context network that combines a CNN with a two-layer transformer for encoding texts, and detecting LLM-generated texts through a sequence tagging task. Research in \cite{wu2023mfd} considers a combination of features such as log likelihood, log rank, entropy, and LLM bias, and by training a neural network classifier, it achieved an average F1 score of 98.41\%. However, a common drawback of these methods is that they all require access to the source model's logits. For other powerful closed-source models where logits are inaccessible, these methods may struggle to be effective.}

\subsubsection{Pre-Training Classifiers}
\paragraph{In-domain Fine-tuning is All You Need}

Within this subsection, we explore methods that involve fine-tuning Transformer-based LMs to discriminate between input texts that are generated by LLMs and those that are not. This approach requires paired samples for the facilitation of supervised training processes. According to \cite{qiu2020pre}, pre-trained LMs have proven to be powerful in natural language understanding, which is crucial for enhancing various tasks in NLP, with text categorization being particularly noteworthy. Esteemed pre-trained models, such as BERT \cite{bert}, Roberta \cite{roberta}, and XLNet \cite{xlnet}, have exhibited superior performance relative to their counterparts in traditional statistical machine learning and deep learning when applied to the text categorization tasks within the GLUE benchmark \cite{glue}.

Moreover, there is an extensive body of prior work \citep{bakhtin2019real, uchendu2020authorship, antoun2023robust, li2023deepfake} that has meticulously examined the capabilities of fine-tuned LMs in detecting LLM-generated text. Notably, studies conducted in 2019 have acknowledged fine-tuned LMs, with Roberta \cite{roberta} being especially prominent, as being amongst the most formidable detectors of LLM-generated text. In the following discourse, we will introduce recent scholarly contributions in this vein, providing an updated review and summary of the methods deployed.

Fine-tuning Roberta provides a robust baseline for detecting text generated by LLMs. \citet{Fagni_2021} observed that fine-tuning Roberta led to optimal classification outcomes in various encoding configurations \cite{gambini2022pushing}, with the subsequent OpenAI detector \citep{radford2019language} also adopting a Roberta fine-tuning approach. Recent works \cite{guo2023close, liu2023argugpt, liu2023check, chen2023gpt, DBLP:journals/corr/abs-2306-07401,DBLP:journals/corr/abs-2306-07401} further corroborated the superior performance of fine-tuned members of the BERT family, such as RoBERTa, in identifying LLM-generated text. On average, these fine-tuned models yielded a 95\% accuracy rate within their respective domains, outperforming zero-shot and watermarking methods, and exhibiting a modicum of resilience to various attack techniques within in-domain settings. Nevertheless, like their counterparts, these encoder-based fine-tuning approaches lack robustness \citep{bakhtin2019real,uchendu2020authorship,antoun2023robust,li2023deepfake}, as they tend to overfit to their training data or the source model’s training distribution, resulting in a decline in performance when faced with cross-domain or unseen data. Additionally, fine-tuning LMs classifiers is limited in facing data generated by different models \cite{sarvazyan2023supervised}. 
\textcolor{black}{Despite this, detectors based on RoBERTa exhibit significant potential for robustness, requiring as few as a few hundred labels to fine-tune and deliver impressive results \cite{DBLP:conf/naacl/RodriguezHGSS22}. mBERT~\cite{DBLP:conf/naacl/DevlinCLT19} has demonstrated consistently robust performance in document-level LLM-generated text classification and various model attribution settings, maintaining optimal performance particularly in English and Spanish tasks. In contrast, encoder models like XLM-RoBERTa~\cite{DBLP:conf/acl/ConneauKGCWGGOZ20} and TinyBERT~\cite{DBLP:conf/emnlp/JiaoYSJCL0L20} have shown significant performance disparities in the same document-level tasks and model attribution setups, suggesting that these two tasks may require different capabilities from the models.}

\paragraph{Contrastive Learning} Data scarcity has propelled the application of contrastive learning \cite{yan2021consert,gao2021simcse,chen2022dual} to LM-based classifiers, with the core of this approach being self-supervised learning. This strategy minimizes the distance between the anchor and positive samples while maximizing the distance to negative samples through spatial transformations. An enhanced contrastive loss, proposed by \citet{liu2022coco}, assigns greater weight to hard-negative samples, thereby optimizing model utility and stimulation to bolster performance in low-resource contexts. This method thoroughly accounts for linguistic characteristics and sentence structures, representing text as a coherence graph to encapsulate its inherent entity consistency. Research findings affirm the potency of incorporating information fact structures to refine LM-based detectors' efficacy, a conclusion echoed by \citet{zhong2020neural}. \citet{conda} proposed ConDA, a contrastive domain adaptation framework, which combines standard domain adaptation technology with the representation capabilities of contrastive learning, greatly improving the model's defense capabilities against unknown models.

\paragraph{Adversarial Learning Methods}
\label{adversarial_learning_methods}
In light of the vulnerability of detectors to different attacks and robustness issues, a significant body of scholarly research has been dedicated to utilizing adversarial learning as a mitigation strategy.
Predominantly, adversarial learning methods bear relevance to fine-tuning LMs methods. Noteworthy recent work by \citet{koike2023outfox} revealed that it is feasible to train adversarially without fine-tuning the model, with context serving as a guide for the parameter-frozen model. We compartmentalize the studies into two categories: Sample Enhancement Based Adversarial Training and Two-Player Games.

A prominent approach within Sample Enhancement Based Adversarial Training centers on deploying adversarial attacks predicated on sample augmentation, with the overarching aim of crafting deceptive inputs to thereby enhance the model's competency in addressing a broader array of scenarios that bear deception potential. 
Specifically, this method emphasizes the importance of sample augmentation and achieves it by injecting predetermined adversarial attacks.
This augmentation process is integral to fortifying the detector's robustness by furnishing it with an expanded pool of adversarial samples. Section 7.2 of the article outlines various potential attack mechanisms, including paraphrase attacks, adversarial attacks, and prompt attacks. \citet{yang2023chatgpt,shi2023red,he2023mgtbench} conducted the adversarial data augmentation process on LLM-generated text, the findings of which indicated that models trained on meticulously augmented data exhibited commendable robustness against potential attacks.

The methods of Two-Player Games fundamentally aligned with the principles underpinning Generative Adversarial Networks \cite{goodfellow2020generative} and Break-It-Fix-It strategies \cite{yasunaga2021break}, typically involve the configuration of an attack model alongside a detection model, with the iterative confrontation between the two culminating in enhanced detection capabilities. \citet{hu2023radar} introduced a framework, RADAR, envisaged for the concurrent training of robust detectors through adversarial learning. This framework facilitates interaction between a paraphrasing model, responsible for generating realistic content that evades detection, and a detector whose goal is to enhance its capability to identify text produced by LLMs. The RADAR framework incrementally refines the paraphrase model, drawing on feedback garnered from the detector and employing PPO \citep{schulman2017proximal}. Despite its commendable performance in countering paraphrase attacks, the study by \citet{hu2023radar} did not provide a comprehensive analysis of RADAR’s defense mechanism against other attack modalities. In a parallel vein, \citet{koike2023outfox} proposed a training methodology for detectors, predicated on a continual interaction between an attacker and a detector. 
Distinct from RADAR, OUTFOX allocates greater emphasis on the likelihood of detectors employing ICL \cite{dong2022survey} for attacker identification. Specifically, the attacker in the OUTFOX framework utilizes predicted labels from the detector as ICL exemplars to generate text that poses detection challenges. Conversely, the detector uses the content generated adversarially as ICL exemplars to enhance its detection capabilities against formidable attackers. This reciprocal consideration of each other's outputs fosters improved robustness in detectors for text generated by LLMs. Empirical evidence attests to the superior performance of the OUTFOX method relative to preceding statistical methods and those based on RoBERTa, particularly in responding to attacks from TF-IDF and DIPPER \cite{krishna2023paraphrasing}.

\textcolor{black}{\paragraph{Features-Enhanced Approaches} In addition to enhancements in training methodology, \citet{DBLP:journals/corr/abs-2304-14106} demonstrated that the extraction of linguistic features can effectively improve the robustness of a RoBERTa-based detector, with benefits observed in various related models. \citet{DBLP:conf/emnlp/CowapGF23} developed an emotion-aware detector by fine-tuning a Pre-trained Language Model (PLM) for sentiment analysis, thereby enhancing the potential of emotion as a signal for identifying synthetic text. They achieved this by further fine-tuning BERT specifically for sentiment classification, resulting in a detection performance F1 score improvement of up to 9.03\%. \citet{DBLP:journals/corr/abs-2309-12934} employed RoBERTa to capture contextual representations, such as semantic and syntactic linguistic features, and integrated Topological Data Analysis to analyze the shape and structure of data, which includes linguistic structure. This approach surpassed the performance of RoBERTa alone on the SynSciPass and M4 datasets. The framework J-Guard~\cite{DBLP:journals/corr/abs-2309-03164} guides existing supervised AI text detectors in detecting AI-generated news by extracting Journalism Features, which help the detector recognize LLM-generated fake news text. This framework exhibits strong robustness, maintaining an average performance decrease as low as 7\% when faced with adversarial attacks.}

\subsubsection{LLMs as Detectors}
\paragraph{Questionable Reliability of Using LLMs}

Several works have examined the feasibility of utilizing LLMs as detectors to discern text generated by either themselves or other LLMs. This approach was first broached by \citet{zellers2019defending}, wherein the text generation model Grover \cite{zellers2019defending} was noted to produce disinformation that was remarkably deceptive due to its inherently controllable nature. Subsequent exploratory analyses by \citet{zellers2019defending} engaging various architectural models like GPT-2 \cite{radford2019language} and BERT \cite{kenton2019bert} revealed that Grover's most effective countermeasure was Grover itself, boasting an accuracy rate of 92\%, while other detector types experienced a decline in accuracy to approximately 70\% as Grover's size increased. 
A recent reevaluation conducted by \citet{bhattacharjee2023fighting} on more recent LLMs like ChatGPT and GPT-4 yielded that neither could reliably identify text generated by various LLMs. 
During the observations, it was noted that ChatGPT and GPT-4 exhibited contrasting tendencies. ChatGPT tended classify text generated by LLMs as if it were written by humans, with a misclassification probability of about 50\%. While GPT-4 leaned towards labeling human-written text as if it were generated by LLMs, and about 95\% of human-written texts are misclassified as LLM-generated texts.
ArguGPT \cite{liu2023argugpt} further attested to the lackluster performance of GPT-4-Turbo in detecting text generated by LLMs, with accuracy rates languishing below 50\% across zero-shot, one-shot, and two-shot settings. These findings collectively demonstrate the diminishing reliability of employing LLMs for direct self-generated text detection, particularly when compared to statistical and neural network methods. This is particularly evident in light of the increasing complexity of LLMs.

\paragraph{ICL: A Powerful Technique for LLM-Based Detection}

Despite the unreliability issues associated with using LLMs for direct detection of LLM-generated text, recent empirical investigations highlight the potential efficacy of ICL in augmenting LLMs' detection capabilities. ICL, a specialized form of cue engineering, integrates examples into cues provided to the model, thereby facilitating the learning of new tasks by LLMs. Through ICL, existing LLMs can adeptly tackle different tasks without necessitating additional fine-tuning. The OUTFOX Detector \cite{koike2023outfox} employs an ICL approach, continuously supplying example samples to LLMs for text generation detection tasks. The experimental findings demonstrate that the ICL strategy outperforms both traditional zero-shot methods and RoBERTa-based detectors.

\subsection{Human-Assisted Methods}

In this section, we delve into human-assisted methods for detecting text generated by LLMs. These methods leverage human prior knowledge and analytical skills, providing notable interpretability and credibility in the detection process.

\subsubsection{Intuitive Indicators}
Several studies have delved into the disparities between human and machine classification capabilities. Human classification primarily depends on visual observation to discern features indicative of text generation by LLMs. \citet{uchendu2022does} noted that a lack of coherence and consistency in LLM-generated text serves as a strong indicator of falsified content. Texts produced by LLMs often exhibit semantic inconsistencies and logical errors. Furthermore, \citet{dugan2023real} identified that the human discernment of LLM-generated text varies across different domains. For instance, LLMs tend to generate more ``generic'' text in the news domain, whereas, in story domains, the text might be more ``irrelevant''. 
\textcolor{black}{\citet{ma2023ai} noted that evaluators of academic writing typically emphasize style. Summaries generated by LLMs often lack detail, particularly in describing the research motivation and methodology, which hampers the provision of fresh insights. In contrast, LLM-generated papers exhibit fewer grammatical and other types of errors and demonstrate a broader variety of expression \cite{yan2023detection,DBLP:journals/corr/abs-2304-11567}. However, these papers commonly use general terms instead of effectively tailored information pertinent to the specific problem context. In human-written texts, such as scientific papers, authors are prone to composing lengthy paragraphs and using ambiguous language~\cite{DBLP:journals/corr/abs-2303-16352}, often incorporating terms like ``but,'' ``however,'' and ``although.''} \citet{dugan2023real} also noted that relying solely on grammatical errors as a detection strategy is unreliable. In addition, LLMs frequently commit factual and common-sense reasoning errors, which, while often overlooked by neural network-based detectors, are easily noticed by humans \cite{jawahar2020automatic}.

\subsubsection{Imperceptible Features}
\citet{ippolito2019automatic} suggested that text perceived as high quality by humans tends to be more easily recognizable by detectors. This observation implies that some features, imperceptible to humans, can be efficiently captured by detection algorithms. 
While humans are adept at identifying errors in many LLM-generated texts, unseen features also significantly inform human decision-making. In contrast, statistical thresholds commonly employed in zero-shot Detector research to distinguish LLM-generated text can be manipulated. However, humans typically possess the ability to detect such manipulations through various metrics,
GLTR \cite{gehrmann2019gltr} pioneered this approach, serving as a visual forensic tool to assist human vetting processes, while also providing rich interpretations easily understandable by non-experts~\cite{DBLP:conf/acl/ClarkASHGS20}.

\subsubsection{Enhancing Human Detection Capabilities}
Recent studies \cite{ippolito2019automatic} indicated that human evaluators might not be as proficient as detection algorithms in recognizing LLM-generated text across various settings. However, exposing evaluators to examples before evaluation enhances their detection capabilities, especially with longer samples. The platform RoFT \cite{dugan2020roft} allows users to engage with LLM-generated text, shedding light on human perception of such text. 
Although revealing true boundaries post-annotation did not lead to an immediate improvement in annotator accuracy, it is worth noting that with proper incentives and motivations, annotators can indeed improve their performance over time \cite{dugan2023real}. 
The SCARECROW framework \cite{dou2021gpt} facilitates the annotation and review of LLM-generated text, outlining ten error types to guide users. The result from SCARECROW reports Manual annotation outperformed detection models on half of the error types, suggesting potential in developing efficient annotation systems despite the associated human overhead.

\subsubsection{Mixed Detection: Understanding and Explanation}
\citet{weng2023towards} introduced a prototype amalgamating human expertise and machine intelligence for visual analysis, premised on the belief that human judgment is the benchmark. Initially, experts label text based on their prior knowledge, elucidating the distinctions between human and LLM-generated text. Subsequently, machine-learning models are trained and iteratively refined based on labeled data. Finally, the most intuitive detector is selected through visual statistical analysis, serving the detection purpose. This granular analysis approach not only bolsters experts’ trust in decision-making models but also fosters learning from the models' behavior to efficiently identify LLM-generated samples.
\section{Evaluation Metrics}
\label{metrics}
Evaluation metrics, indispensable for the assessment of model performance within any NLP task, warrant meticulous consideration. In this section, we enumerate and discuss metrics conventionally utilized in the tasks of LLM-generated text detection. These metrics include Accuracy, Paired Accuracy, Unpaired Accuracy, Recall, Human-written Recall (HumanRec), LLM-generated Recall (LLMRec), Average Recall (AvgRec), F1 Score, and Area Under the Receiver Operating Characteristic Curve (AUROC). Furthermore, we discuss the advantages and drawbacks associated with each metric to facilitate informed metric selection for varied research scenarios in subsequent studies.

The confusion matrix can help effectively evaluate the performance of the classification task and describes all possible results (four types in total) of the LLM-generated text detection task:
\begin{itemize}
\item \textit{True Positive (TP)} refers to the result of the positive category (LLM-generate text) correctly classified by the model.
\item \textit{True Negative (TN)} refers to the result of the negative category (human-written text) correctly classified by the model.
\item \textit{False Positive (FP)} refers to the result of the positive category (LLM-generate text) incorrectly classified by the model.
\item \textit{False Negative (FN)} refers to the result of the negative category (human-written text) incorrectly predicted by the model.
\end{itemize}
The evaluation metrics introduced below can all be described by TP, TN, FP, and FP.

\paragraph{Accuracy} Accuracy serves as a general metric, denoting the ratio of correctly classified texts to the total text count. While suitable for balanced datasets, its utility diminishes for unbalanced ones due to sensitivity to category imbalance. The metrics of Paired and Unpaired Accuracy have also found application in \cite{zellers2019defending,zhong2020neural} to evaluate the detector's ability in different scenarios. In the unpaired setting, the discriminator must independently classify each test sample as human or machine. In the paired setting, the model is given two test samples with the same metadata, one real and one generated by the large model. The discriminator must assign a higher machine probability to articles written by large models than to articles written by humans. These indicators are used to measure the performance of the algorithm on data in different scenarios. Relatively speaking, the detection difficulty of unpaired settings is higher than that of paired settings. Accuracy can be described by the following formula:
\begin{small} 
\begin{equation}
\begin{aligned}
    Accuracy
    &=\frac{\text{correctly detected samples}}{\text{all samples}}\\ 
    &=\frac{TP+TN}{TP+TN+FP+FN} 
\end{aligned}
\end{equation}
\end{small}

\paragraph{Precision} Precision is a measure of the correctness of real predictions and refers to the proportion of correctly detected LLM-generated samples among all detected LLM-generated samples. This metric is very useful in situations where we are concerned about false positives. When a sample is not LLM-generated, but is classified as LLM-generated text, this erroneous result may reduce the user's impression of the model or even cause the negative impact on business. Therefore, improving precision is also important in the LLM-generated text detection task. This Metric can be described by the following formula:
\begin{small} 
\begin{equation}
\begin{aligned}
    Precision
    &=\frac{\text{correctly detected LLM-generated samples}}{\text{all detected LLM-generated samples}}\\ 
    &=\frac{TP}{TP+FP} 
\end{aligned}
\end{equation}
\end{small}

\paragraph{Recall} Recall represents the proportion of actual machine-generated texts accurately identified as such. This metric is invaluable in contexts where underreporting must be minimized, as in instances requiring the capture of the majority of machine-generated texts. AvgRec, the mean recall across categories, is particularly useful for multi-category tasks requiring collective performance assessment across categories. HumanRec and LLMRec denote the proportions of texts accurately classified as human-written and machine-generated, respectively, shedding light on the model's differential performance on these two classes. Recall, HumanRec, LLMRec, and AvgRec can be described by the following formulas respectively:
\begin{small} 
\begin{equation}
Recall=\frac{TP}{TP+FN} 
\end{equation}
\end{small}
\begin{small} 
\begin{equation}
HumanRecall=\frac{\text{correctly detected human-written samples}}{\text{all human-written samples}}
\end{equation}
\end{small}
\begin{small} 
\begin{equation}
LLMRecall=\frac{\text{correctly detected LLM-generated samples}}{\text{all LLM-generated samples}}
\end{equation}
\end{small}
\begin{small} 
\begin{equation}
AvgRecall=\frac{HumanRecall+LLMRecall}{2}
\end{equation}
\end{small}

\paragraph{False Positive Rate (FPR)} The FPR refers to the proportion of all actual human-written samples that are incorrectly detected as LLM-generated samples. This metric can measure the proportion of incorrect predictions made by the model in samples that are actually written by humans. It helps to understand the false positive rate of the model and thus has a higher sensitivity for the detection of LLM-generated samples. This metric can be described by the following formula:

\begin{small} 
\begin{equation}
\begin{aligned}
    FPR
    &=\frac{\text{incorrectly detected LLM-generated samples}}{\text{all human-written samples}}\\ 
    &=\frac{FP}{FP+TP} 
\end{aligned}
\end{equation}
\end{small}

\paragraph{True Negative Rate (TNR)} The TNR refers to the proportion of samples that are correctly detected as human-written among all actual human-written samples. This metric measures how accurately the model predicts human-written samples, but does not take into account the FPR, where text that is actually human-written is incorrectly detected as LLM-generated text. This metric can be described by the following formula:
\begin{small} 
\begin{equation}
\begin{aligned}
    TNR
    &=\frac{\text{correctly detected human-written samples}}{\text{all human-written samples}}\\ 
    &=\frac{TN}{TN+FP} 
\end{aligned}
\end{equation}
\end{small}

\paragraph{False Negative Rate (FNR)} The FNR refers to the proportion of all actual LLM-generated samples that are incorrectly detected as human-written. This metric helps understand how misinterpreted the model is for LLM-generated text. This metric can be described by the following formula:
\begin{small} 
\begin{equation}
    \begin{aligned}
        FNR
        &=\frac{\text{incorrectly detected human-written samples}}{\text{all LLM-generated samples}}\\ 
        &=\frac{FN}{FN+TP} 
    \end{aligned}
\end{equation}
\end{small}

\paragraph{$F_1$ Score} The $F_1$ Score constitutes a harmonic mean of precision and recall, integrating considerations of false positives and false negatives. It emerges as a prudent choice when a balance between precision and recall is imperative. The $F_1$ score can be calculated using the following formula:
\begin{small} 
\begin{equation}
\begin{aligned}
    F_1=
    &2*\frac{Precision*Recall}{Precision+Recall}\\
    &=\frac{2TP}{2TP+FP+FN} 
\end{aligned}
\end{equation}
\end{small}

\paragraph{AUROC} The AUROC metric, derived from Receiver Operating Characteristic curves, considers true and false positive rates at varying classification thresholds, proving beneficial for evaluating classification efficacy at different thresholds. This is particularly crucial in scenarios necessitating specific false positive and miss rates, especially within the context of unbalanced datasets and binary classification tasks. Given that the detection rate of zero-shot detection methods significantly hinges on threshold values, the AUROC metric is commonly employed to evaluate their performance across all possible thresholds. The calculation formula of AUROC is as follows:
\begin{small} 
\begin{equation}
AUROC = \int_{0}^{1} \frac{TP}{TP+FP} \mathrm{d} \frac{FP}{FP+TN} 
\end{equation}
\end{small}               
\section{Important Issues of LLM-generated Text Detection}
\label{issues}
In this section, we discuss the main issues and limitations of contemporary SOTA techniques designed for detecting text generated by LLMs. It is important to note that no technique has been acknowledged as infallible. The issues elucidated herein may pertain specifically to one or multiple classes of detectors.

\subsection{Out of Distribution Challenges}
Out-of-distribution issues significantly impede the efficacy of current techniques dedicated to the detection of LLM-generated text. This section elucidates the constraints of these detectors to variations in domains and languages.

\paragraph{Cross-domain}
The dilemma of cross-domain application is a ubiquitous challenge inherent to numerous NLP tasks. Studies conducted by \citet{antoun2023robust,li2023deepfake} underscored considerable limitations in the performance of sophisticated detectors, including but not limited to DetectGPT \cite{mitchell2023detectgpt}, GLTR \cite{gehrmann2019gltr}, and fine-tuned Roberta methods when applied to cross-domain data. These detectors exhibit substantial performance degradation when confronted with out-of-distribution data prevalent in real-world scenarios, with the efficacy of some classifiers marginally surpassing that of random classification. This disparity between high reported performance and actual reliability underlines the need for critical evaluation and enhancement of existing methods.

\paragraph{Cross-lingual}
The issue of cross-lingual application introduces a set of complex challenges that hinder the global applicability of existing detector research. Predominantly, contemporary detectors designed for LLM-generated text primarily target monolingual applications, often neglecting to evaluate and optimize performance across multiple languages. \citet{wang2023m4} and \citet{chaka2023detecting} noted the lack of control observed in multilingual LLM-generated text detectors across various languages, despite the existence of certain language migration capabilities. We emphatically draw attention to these cross-lingual challenges as addressing them is pivotal for enhancing the usability and fairness of detectors for LLM-generated text. Moreover, recent research \cite{liang2023gpt} revealed a discernible decline in the performance of state-of-the-art detectors when processing texts authored by non-native English speakers. Although employing effective prompt strategies can alleviate this bias, it also inadvertently allows the generated text to bypass the detectors. Consequently, there is a risk that detectors might inadvertently penalize writers who exhibit non-standard linguistic styles or employ limited expressions, thereby introducing issues of discrimination within the detection process.

\textcolor{black}{
\paragraph{Cross-llms}
Another significant out-of-distribution issue in the LLM-generated text detection task is the cross-llms challenge. Current white-box detection approaches primarily rely on accessing the source model and comparing features such as Log-likelihood. Consequently, white-box methods may underperform when encountering text generated by unknown LLMs. The results of DetectGPT \cite{mitchell2023detectgpt} highlight the vulnerability of white-box methods when dealing to unknown models, particularly when encountering powerful models like GPT-3.5-Turbo. However, the recent findings from Fast-DetectGPT \cite{bao2023fast} show that statistical comparisons with surrogate models can significantly mitigate this issue. Additionally, identifying the type of the generative model before applying white-box methods could be beneficial. In this regard, the methodologies of Siniff \cite{DBLP:journals/corr/abs-2304-14072}, SeqXGPT~\cite{DBLP:journals/corr/abs-2310-08903}, and LLMDet~\cite{DBLP:conf/emnlp/WuPSCC23} may provide useful insights. On the other hand, methods based on neural classifiers, especially those fine-tuned classifiers susceptible to overfitting training data, may struggle to recognize types of LLMs not seen during training. Thus, for newly emerging LLMs, detectors may not effectively identify them~\cite{DBLP:conf/coling/PagnoniGT22}. For instance, the OpenAI detector\footnote{\url{openai-community/roberta-large-openai-detector}} (trained on texts generated by GPT-2) struggles to discern texts generated by GPT-3.5-Turbo and GPT-4,achieving an AUROC of only 74.74\%, while itperforms nearly perfectly on GPT-2 generated texts~\cite{bao2023fast}. The results of \cite{DBLP:conf/clef/SarvazyanGRF23} demonstrate that supervised LLM-generated text detectors exhibit good generalization capabilities across model scales but have limitations in generalizing across model families. Enhancing the cross-llms robustness of neural classifiers is thus essential for the practical deployment of detectors. Nonetheless, classifiers fine-tuned on Roberta still possess strong transfer capabilities, and with additional fine-tuning on just a few hundred samples, detectors can effectively generalize to texts generated by other models. Therefore, incorporating LLM-generated text from various sources into the training data could substantially improve the cross-llms robustness of detectors in real-world applications, even with a small sample size.
}

\subsection{Potential Attacks}
\label{potential_attacks}
Potential attacks significantly contribute to the ongoing unreliability of current LLM-generated text detectors. We present the current effective attacks to push researchers to focus on more comprehensive defensive measures.

\paragraph{Paraphrase Attacks}

Paraphrasing attacks are one of the most effective attacks that can be fully effective against detectors using watermarking technology as well as fine-tuned supervised detectors and zero-shot detectors \citep{sadasivan2023can, orenstrakh2023detecting}. 
The underlying principle involves applying a lightweight paraphrase model on LLMs' outputs and changing the distribution of lexical and syntactic features of the text by paraphrasing, thereby confusing the detector. \citet{sadasivan2023can} reported on Parrot \citep{prithivida2021parrot}, a T5-based paraphrase model and DIPPER \citep{krishna2023paraphrasing}, an 11B paraphrasing model that allows for tuning paraphrase diversity and the degree of content reordering that attacks the overall superiority of existing detection methods. Although retrieval-based approaches have been shown to defend effectively against paraphrasing attacks \citep{krishna2023paraphrasing}, implementing such defenses requires ongoing maintenance by the language model API provider and is still susceptible to recursive paraphrasing attacks \citep{sadasivan2023can}.

\paragraph{Adversarial Attacks}

\textcolor{black}{Normal LLM-generated texts are highly identifiable, yet adversarial perturbations, such as substitution, can effectively reduce the accuracy of detectors \cite{DBLP:journals/corr/abs-2402-00412}.} We summarise attacks that process on textual features as adversarial attacks, including cutoff (cropping a portion of the feature or input) \cite{shen2020simple}, shuffle (randomly disrupting the word order of the input) \cite{lee2020slm}, mutation (character and word mutation) \cite{liang2023mutation}, word swapping (substituting other suitable words given the context) \cite{shi2020robustness,ren2019generating,DBLP:conf/ijcnn/CrothersJVB22} and misspelling \cite{gao2018black}. There are also adversarial attack frameworks such as TextAttack~\cite{DBLP:conf/emnlp/MorrisLYGJQ20}, which can build an attack from four components: an objective function, a set of constraints, a transformation, and a search method. \citet{shi2023red} and \citet{he2023mgtbench} reported on the effectiveness of the permutation approach on attack detectors. Specifically, \citet{shi2023red} replaced words with synonyms based on context, which forms an effective attack on the fine-tuned classifier, watermarking \cite{kirchenbauer2023watermark}, and DetectGPT \cite{mitchell2023detectgpt}, reducing detector performance by more than 18\%, 10\%, and 25\% respectively. \citet{he2023mgtbench} employed probability-weighted word saliency \cite{ren2019generating} to generate adversarial examples, which further maintains semantic similarity. 

\textcolor{black}{\citet{DBLP:journals/ijdsa/StiffJ22} utilized the DeepWordBug~\cite{DBLP:conf/sp/GaoLSQ18} adversarial attack algorithm to introduce character-level perturbations to generated texts, including adjacent character swaps, character substitutions, deletions, and insertions, which resulted in more than a halving of the performance of the OpenAI large detector.\footnote{\url{openai-community/roberta-large-openai-detector}} \citet{DBLP:journals/corr/abs-2002-11768} presented two types of black-box attacks against these detectors: random substitutions of characters with visually similar homoglyphs and the intentional misspelling of words. These attacks drastically reduced the recall rate of popular neural text detectors from 97.44\% to 0.26\% and 22.68\%, respectively. Moreover, \citet{DBLP:conf/acl-insights/BhatP20} showed that detectors are more sensitive to syntactic perturbations, including breaking longer sentences, removing definite articles, using semantic-preserving rule conversions (such as changing ``that’s'' to ``that is''), and reformatting paragraphs of machine-generated text.}

\textcolor{black}{Although existing detection methods are highly sensitive to adversarial attacks, different types of detectors exhibit varying degrees of resilience to such attacks. \citet{DBLP:journals/corr/abs-2306-05871} reported that supervised approaches are effective defensive measures against these attacks: training on adversarial samples can significantly improve a detector's ability to recognize texts that have been manipulated by such attacks. Additionally, \citet{DBLP:journals/corr/abs-2302-00509} explored the impact of semantic perturbations on the Grover detector, finding that synonym substitution, fake-fake replacement, insertion instead of substitution, and changes in the position of substitution had no effect on Grover's detection capabilities. However, adversarial embedding techniques can effectively deceive Grover into classifying false articles as genuine. The attack degrades the performance of the fine-tuning classifier significantly, even though the distributional features of the attack can be learned by the fine-tuning classifier to form a strong defense. }

\paragraph{Prompt Attacks}
Prompt attacks pose a significant challenge for current LLM-generated text detection techniques. The quality of LLM-generated text is associated with the complexity of the prompts that instruct LLMs to generate text. As the model and corpus size increase, LLMs emerge with excellent ICL capabilities for more complex text generation capabilities. Numerous efficient prompting methods have been developed, including few-shot prompt~\citep{NEURIPS2020_1457c0d6}, combining prompt \citep{zhao2021calibrate}, Chain of Thought (CoT) \citep{wei2022chain}, and zero-shot CoT \citep{kojima2023large}, etc., which significantly enhance the quality and capabilities of LLMs. Existing works on LLM-generated text detectors primarily utilize datasets created with simple direct prompts. For instance, the study by \citet{guo2023close} demonstrates that detectors might struggle to identify text generated with complex prompts. \citet{liu2023check} reported a noticeable decrease in the detection ability of a detector using a fine-tuned language model when faced with varied prompts, which indicates that the use of different prompts results in large differences in the detection performance of existing detectors \cite{DBLP:journals/corr/abs-2311-08369}.

The Substitution-based Contextual Example Optimisation method, as proposed by \citet{lu2023large}, employs sophisticated prompts to bypass the defenses of current detection systems. This leads to an appreciable reduction in the Area Under the Curve (AUC), averaging a decrease of 0.54, and achieves a higher success rate with better text quality compared to paraphrase attacks. It is worth mentioning that both paraphrase attacks and adversarial attacks mentioned above could be executed through careful prompt design~\cite{shi2023red,koike2023outfox}. With ongoing research in prompt engineering, the risk posed by prompt attacks is expected to escalate further. This underscores the need for developing more robust detection methods that can effectively counteract such evolving threats.

\textcolor{black}{
\paragraph{Training Threat Models}
Further training of language models has been preliminarily proven to effectively attack existing detectors. \citet{nicks2023language} used the ``humanity'' scores of various open source and commercial detectors as a reward function for reinforcement learning, which fine-tunes language models to confound existing detectors. Without significantly altering the model, further fine-tuning of the Llama-2-7B can reduce the AUROC of the OpenAI RoBERTa-Large detector from 0.84 AUROC to 0.62 AUROC in a short training period. A similar idea is demonstrated in \cite{DBLP:journals/corr/abs-2310-16992}: using reinforcement learning to refine generative models can successfully circumvent BERT-based classifiers with detection accuracy as low as 0.15 AUROC, even when using linguistic features as a reward function. \citet{DBLP:conf/emnlp/KumarageSMG023} proposes a universal evasion framework named EScaPe to guide PLMs in generating ``human-like text'' that may mislead detectors. Through evasive soft prompt learning and transfer, the performance of DetectGPT and OpenAI Detector can be effectively reduced by up to 40\% AUROC. The results from \cite{DBLP:journals/corr/abs-2304-08968} reveal another potential vulnerability of detectors. If a generative model can access the human-written text used to train the detector and use them for fine-tuning, it is impossible to use detector for text detection on this generative model. This indicate that LLMs trained on more human-written corpus will be more robust against existing detectors, and training against a specific detector can provide the LLMs with a sharp spear to breach its defenses.
}

\subsection{Real-World Data Issues}
\textcolor{black}{
\paragraph{Detection for Not Purely LLM-generated Text} 
\label{detection_for_not_purely_llm_generated_text}
In practice, there are many texts that are not purely generated by LLMs, and they may even contain a mix of human-written text. Specifically, this can be categorized as either data-mixed text or human-edited text. Data-mixed text refers to the sentence or paragraph level mixture of human-written text and LLM-generated text. For instance, in a document, some sentences may be generated by LLMs, while others are written by humans. In such cases, identifying the category of the document becomes challenging. Data-mixed text necessitates more fine-grained detection methods, such as sentence-level detection, to effectively address this challenge. However, current LLM-generated text detectors struggle to perform effectively with short texts. Recent research, such as that by \citet{DBLP:journals/corr/abs-2310-08903}, indicates that sentence-level detection appears to be feasible. Furthermore, we are very pleased to observe that studies have started to propose and attempt to solve this issue. \citet{zeng2023towards} proposed a two-step method to effectively identify a mix of human-written and LLM-generated text. This method first uses contrastive learning to distinguish between content generated by LLMs and human-written content. It then calculates the similarity between adjacent prototypes, assuming that a boundary exists between the least similar adjacent prototypes.}

\textcolor{black}{
Another issue that has not been fully discussed is the human-edited text. For example, after applying LLM to generate a text, humans often edit and modify certain words or passages. The detection of such text poses a significant challenge and is an issue we must confront, as it is prevalent in real-world applications. Therefore, there is an urgent need to organize relevant datasets and define tasks to address this issue. One potential approach for tackling this problem is informed by experimental results from paraphrasing and adversarial perturbation attacks. These methods effectively simulate how individuals might use LLMs to refine text or make word substitutions. However, tend to degrade in performance when dealing with paraphrased text, current mainstream detectors tend to degrade in performance when dealing with paraphrased text \cite{DBLP:journals/corr/abs-2002-11768}, although certain black-box detectors display relatively good robustness. Another potential solution could involve breaking down the detection task to the word level, but as of now, there is no research directly addressing this.
}

\paragraph{Data Ambiguity}
Data ambiguity remains a challenge in the field of LLM-generated text detection, with close ties to the inherent mechanics of the detection technology itself. The pervasive deployment of LLMs across various domains exacerbates this issue, rendering it increasingly challenging to discern whether training data comprises human-written or LLM-generated text. Utilizing LLM-generated text as training data under the misapprehension that it is human-written inadvertently instigates a detrimental cycle. Within this cycle, detectors, consequently trained, demonstrate diminished efficacy in distinguishing between human-written and LLM-generated text, thereby undermining the foundational premises of detector research. It is imperative to acknowledge that this quandary poses a significant, pervasive threat to all facets of detection research, yet, to our knowledge, no existing studies formally address this concern. An additional potential risk was articulated by \citet{alemohammad2023selfconsuming}, who posited that data ambiguity might precipitate the recycling of LLM-generated data in the training processes of subsequent models. This scenario could adversely impact the text generation quality of these emergent LLMs, thereby destabilizing the research landscape dedicated to the detection of LLM-generated text.

\textcolor{black}{
\subsection{Impact of Model Size on Detectors}
Many researchers are concerned about the impact of the model size on detectors, which can be viewed from two perspectives: one is the size of the generative model, and the other is the size of the supervised classifiers.
The size of the generative model is closely related to the quality of the generated text. Generally speaking, texts generated by smaller-sized models are easier to recognize, while those generated by larger models pose a greater challenge for detection. Another issue of concern is how the texts generated by models of different sizes affect the detectors when used as training samples. \citet{pu2023zero} report that detectors trained with data generated by medium-sized LLMs can generalize to larger versions without any samples, while training samples generated by overly large or small models may reduce the generalization ability of the detectors. \citet{DBLP:journals/corr/abs-2309-13322} further explores the apparent negative correlation between classifier effectiveness and the size of the generative model. The results show that text generated by larger LLMs is more difficult to detect, especially when the classifier is trained on data generated by smaller LLMs. Aligning the distribution of the generative models for the training and test sets can improve the performance of the detectors.
From the perspective of the size of the supervised classifiers, the detection capability of the detectors is directly proportional to the size of the fine-tuned LMs \cite{guo2023close}. However, recent findings suggest that while larger detectors perform better on test sets with the same distribution as the training set, their generalization ability is somewhat diminished.
}

\subsection{Lack of Effective Evaluation Framework}
\label{lack_of_effective_evaluation_framework}

\textcolor{black}{
A widespread phenomenon is that many studies claim their detectors exhibit impressive and robust performance. However, in practical experiments, these methods often perform less than satisfactorily on the test sets created by other researchers. This variance is due to researchers using different strategies to construct their test sets. Variables such as the parameters used to generate the test set, the computational environment, text distribution, and text processing strategies, including truncation, can all influence the effectiveness of detectors. Due to these factors' complex nature, the reproducibility of evaluation results is often compromised, even when researchers adhere to identical dataset production protocols. We elaborate on the limitations of existing benchmarks in \autoref{data}, where we advocate for the creation of a high-quality and comprehensive evaluation framework. We encourage future research to actively implement these frameworks to maintain consistency in testing standards. Furthermore, we call upon researchers focusing on specific issues to openly share their test sets, emphasizing the strong adaptability of current evaluation frameworks to integrate them. In conclusion, setting an objective and fair benchmark for detector comparison is essential to propel research in detecting LLM-generated text forward, rather than persisting in siloed efforts.
}
\section{Future Research Directions}
\label{future_research_directions}
In this section, we explore potential directions for future research aimed at better construction of more efficient and realistically effective detectors.

\subsection{Building Robust Detectors with Attacks}
The attack methods introduced in \autoref{potential_attacks}, encompass Paraphrase Attacks \citep{sadasivan2023can}, Adversarial Attacks \citep{he2023mgtbench}, and Prompt Attacks \citep{lu2023large}. These methods underscore the primary challenges impeding the utility of current detectors. While recent research, such as \citet{yang2023chatgpt}, has addressed robustness against specific attacks, it often neglects potential threats posed by other attack forms. Consequently, it is imperative to develop and validate diverse attack types, thereby gaining insights into vulnerabilities inherent to LLM-generated text detectors. We further advocate for the establishment of comprehensive benchmarks to assess existing detection strategies. Although some studies \citep{he2023mgtbench,wang2023m4} purport to provide such benchmarks, the scope and diversity of the validated attacks remain limited.

\subsection{Enhancing the Efficacy of Zero-shot Detectors}

Zero-shot methods stand out as notably stable detectors \cite{deng2023efficient}. Crucially, they offer enhanced controllability and interpretability for users \cite{mitrovic2023chatgpt}. Recent research \citep{giorgi2023slept,liao2023differentiate} has elucidated distinct disparities between LLM-generated text and human-written text, underscoring a tangible and discernible gap between the two. This revelation has invigorated research in the domain of LLM-generated text detection. We advocate for a proliferation of studies that delve into the nuanced distinctions between LLM-generated texts and human-written text, spanning from low-dimensional to high-dimensional features. Unearthing metrics that more accurately distinguish the two can bolster the evolution of automatic detectors and furnish more compelling justifications for decision-making processes. \textcolor{black}{We have observed that the latest emerging black-box zero-shot methods \cite{DBLP:journals/corr/abs-2305-17359,DBLP:journals/corr/abs-2401-12970,DBLP:conf/emnlp/ZhuYCCFHDL0023,DBLP:conf/bea/QuidwaiLD23,DBLP:journals/corr/abs-2311-07700} demonstrate enhanced stability and application potential compared to white-box based zero-shot methods by extracting discriminative metrics that are independent of white-box models. These methods do not rely on an understanding of the model's internal workings, thereby offering broader applicability across various models and environments.}

\subsection{Optimizing Detectors for Low-resource Environments}

Many contemporary detection techniques tend to overlook the challenges faced by resource-constrained settings, neglecting the need for resources in developing the detector. The relative efficacy of various detectors across different data volume settings remains inadequately explored. Concurrently, determining the minimal resource prerequisites for different detection methods to yield satisfactory results is imperative. Beyond examining the model's adaptability across distinct domains \cite{rodriguez2022cross} and languages \cite{wang2023m4}, we advocate for investigating the defensive adaptability against varied attack strategies. Such exploration can guide users in selecting the most beneficial approach to establish a dependable detector under resource constraints.

\subsection{Detection for Not Purely LLM-Generated Text}

\textcolor{black}{
In \autoref{detection_for_not_purely_llm_generated_text}, we highlight a significant challenge encountered in real-world scenarios: the detection of text that is not purely produced by LLMs. We examine this issue by separately discussing texts that are a mixture of data sources and those that have been edited by humans, and review the latest related work and propose potential solutions, which are still pending verification. We emphasize that organizing relevant datasets and defining tasks to address this issue is an urgent need at present, because fundamentally, this type of text may be the most commonly encountered in detector applications.
}

\subsection{Constructing Detectors Amidst Data Ambiguity}

A significant challenge that arises is verifying the authenticity of training data. When aggregating textual data from sources such as blogs and web comments, there is a potential risk of inadvertently including a substantial amount of LLM-generated text. This incorporation can fundamentally compromise the integrity of detector research, perpetuating a detrimental feedback loop. We urge forthcoming detection studies to prioritize the authenticity assessment of real-world data, anticipating this as a pressing challenge in the future.

\subsection{Developing Effective Evaluation Framework Aligned With Real-World}

\textcolor{black}{In \autoref{lack_of_effective_evaluation_framework}, we analyze the objective differences between evaluation environments and real-world settings, which limit the effectiveness of existing detectors when applied in practice. On one hand, there may be biases in the construction of test sets in many works  because they often favor the detectors built by their creators; on the other hand, current benchmarks frequently reflect idealized scenarios far removed from real-world applications. We call on researchers to develop a fair and effective evaluation framework closely linked to the practical needs of LLM-generated detection tasks. For instance, considering the necessity of the application domain, the black-box nature of LLM-generated texts, and the various attacks and post-editing strategies that texts may encounter. We believe such an evaluation framework will promote the research and development of detectors that are more practical and aligned with real-world scenarios.
}

\subsection{Constructing Detectors with Misinformation Discrimination Capabilities}

Contemporary detection methodologies have largely overlooked the capacity to discern misinformation. Existing detectors primarily emphasize the distribution of features within text generated by LLMs, while often overlooking their potential for factual verification. A proficient detector should possess the capability to discern the veracity or falsity of factual claims presented in text. In the initial stages of generative modeling's emergence, when it had yet to pose significant societal challenges, the emphasis was on assessing the truth or falsity of the content in LLM-generated text, with less regard for its source \cite{schuster2020limitations}. Constructing detectors with misinformation discrimination capabilities can aid in more accurately attributing the source of text, rather than relying solely on distributional features, and subsequently contribute to mitigating the proliferation of misinformation. Recent studies \cite{gao2023rarr,chern2023factool} highlight the potential of LLMs to detect factual content in texts. We recommend bolstering such endeavors through integration with external knowledge bases \citep{asai2023retrieval} or search engines \cite{liang2023taskmatrix}.
\section{Conclusion}
With the widespread development and application of LLMs, the pervasive presence of LLM-generated text in our daily lives has transitioned from expectation to reality. 
LLM-generated text detectors play a pivotal role in distinguishing between human-written and LLM-generated text, serving as a crucial defense against the misuse of LLMs for generating deceptive news, engaging in scams, or exacerbating issues such as educational inequality. In this survey, we introduce the task of LLM-generated text detection, outline the sources contributing to enhanced LLM-generated text capabilities, and highlight the escalating demand for efficient detectors. 
We also list datasets that are popular or promising, pointing out the challenges and requirements associated with existing detectors. In addition, we shed light on the critical limitations of contemporary detectors, including issues related to out-of-distribution data, potential attacks, real-world data issues, and the lack of an effective evaluation framework, to direct researchers' attention to the focal points of the field, thereby sparking innovative ideas and approaches. Finally, we propose potential future research directions that are poised to guide the development of more powerful and effective detection systems.

\begin{acknowledgments}
This work was supported in part by the Major Program of the State Commission of Science Technology of China (Grant No. 2020AAA0106701), the Science and Technology Development Fund, Macau SAR (Grant Nos. FDCT/0070/2022/AMJ, FDCT/060/2022/AFJ) and the Multi-year Research Grant from the University of Macau (Grant No. MYRG-GRG2023-00006-FST-UMDF).
\end{acknowledgments}

\starttwocolumn
\bibliography{bib_ready}

\end{document}